\documentclass[runningheads]{llncs}

\usepackage{eccv}

\usepackage{eccvabbrv}

\usepackage{graphicx}
\usepackage{booktabs}
\usepackage{bm}
\usepackage{amsmath}
\usepackage{multirow}
\usepackage{booktabs}

\usepackage[accsupp]{axessibility}  

\usepackage{hyperref}

\usepackage{orcidlink}

\begin{document}

\title{CapFrame: Text-Instructed Viewpoint Grounding in 3D Gaussian Scenes via Geometric Pseudo Labels} 

\titlerunning{CapFrame}

\author{Jirong Li\inst{1}\orcidlink{0009-0001-2858-7830} \and Satoshi Ikehata\inst{2,3}\orcidlink{0000-0002-6061-7956} \and Shuhei Kurita\inst{1,3}\orcidlink{0000-0001-7415-3120} \and Ikuro Sato\inst{1,2}\orcidlink{0000-0001-5234-3177}}

\authorrunning{J.~Li et al.}

\institute{
Institute of Science Tokyo, Japan
\and
Denso IT Laboratory, Inc., Japan
\and
National Institute of Informatics, Japan\\
\email{jirong\_li@d-itlab.c.titech.ac.jp}
}

\maketitle
\begin{abstract}
3D Gaussian Splatting (3DGS) enables photorealistic real-time novel view synthesis, yet placing a virtual camera to capture a desired frame remains largely manual. Existing language-guided approaches in 3D scenes mainly focus on \emph{object-centric grounding}, determining \textit{what} to observe but rarely controlling \textit{how} it should appear in a single frame, such as subject orientation or frame layout. To address this limitation, we introduce a new task, Text-Instructed Viewpoint Grounding (TIVG), which aims to identify a 6-DoF camera pose in a 3D Gaussian scene whose rendered frame aligns with a text instruction. To solve this task, we propose {\it CapFrame}, a partially differentiable framework that converts language into \emph{geometric pseudo labels} for camera pose optimization. CapFrame follows a {\it Retrieve--Translate--Refine} pipeline: it retrieves relevant views and ranks them through a \emph{Question-Evaluation} process with MLLMs, translates the instruction into orientation and layout pseudo labels, and refines the camera pose via differentiable optimization with layout and orientation losses in 3DGS. Experiments on 38 real-world scenes with 135 instructions indicate that CapFrame produces viewpoints better aligned with texts than heuristic viewpoint search and adapted trajectory generation baselines, validated by VLM metrics, MLLM judges, and user studies. Code is available at: \url{https://github.com/jirongli/CapFrame}
\end{abstract}

\section{Introduction}
\begin{figure}[tb]
  \centering
  \includegraphics[width=\linewidth]{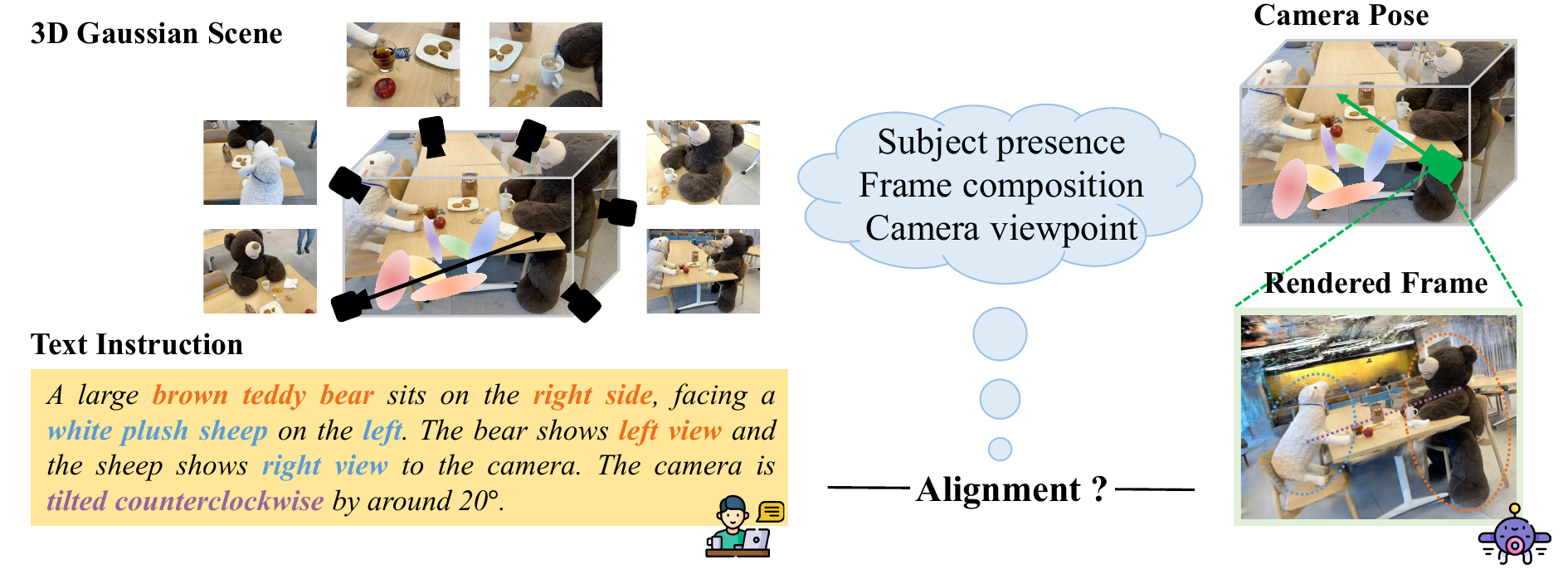}
  \caption{\textbf{Text-instructed viewpoint grounding in a 3DGS scene.} Given a 3D Gaussian scene and a text instruction, the goal is to identify a camera pose whose rendered frame aligns with the described subject, composition and viewpoint.
  }
  \label{fig:task}
\end{figure}
Recent advances in 3D representation technologies such as Neural Radiance Fields (NeRF)~\cite{mildenhall2021nerf} and 3D Gaussian Splatting (3DGS)~\cite{kerbl20233d}, enable photorealistic novel view synthesis for applications spanning VR/AR~\cite{jiang2024vr, zhai2025splatloc}, content creation~\cite{yi2024gaussiandreamer, zhou2024dreamscene360}, and scene editing~\cite{chen2024gaussianeditor, wen2025intergsedit}. Notably, 3DGS has gained prominence due to its real-time rendering and high visual fidelity, making it well-suited for interactive environments. However, despite its efficient rendering, identifying desirable viewpoints still requires tedious manual trial-and-error and scales poorly as scenes grow larger and more complex. Therefore, intent-aligned automatic viewpoint recommendation is crucial for intuitive scene interaction.

Recent language-guided 3D methods~\cite{qin2024langsplat, wu2024opengaussian, shi2024language, li2025langsplatv2} excel at semantic object localization and trajectory generation~\cite{courant2024exceptional, li2024director3d, liu2024chatcam, zhang2025gendop, liu2024splatraj}, but primarily focus on \textit{object-centric grounding}. They identify ``what'' to observe but overlook ``how'' it should be presented. Real-world instructions are inherently compositional, specifying orientation, spatial relationships, or photographic attributes (\eg, ``\textit{a close-up photo of a cat from the front}''). Existing work ensures object presence but fails to enforce such intentional framing constraints. While Multimodal Large Language Models (MLLMs)~\cite{bai2025qwen3, grattafiori2024llama, liu2024llavanext, yang2023dawn} offer strong zero-shot spatial reasoning to interpret such constraints, translating instruction text into concrete 3DGS viewpoints remains unexplored.

In this work, we formalize a new task, {\it Text-Instructed Viewpoint Grounding} for 3DGS, which aims to identify a 6-DoF camera pose that captures a frame aligned with a text instruction. To accomplish this, we propose \textbf{CapFrame}, a partially differentiable framework for text-instructed camera placement in 3D Gaussian scenes. CapFrame follows a three-stage \textbf{Retrieve--Translate--Refine} pipeline. In the \textbf{Retrieve} stage, we retrieve semantically relevant views from the training images and perform fine-grained ranking through a \textit{Question-Evaluation} (QE) process guided by an MLLM, providing initialization for camera pose optimization. 
In the \textbf{Translate} stage, we convert compositional language into geometric pseudo labels, including subject--Gaussian associations, orientation pseudo labels, and layout pseudo labels. 
In the \textbf{Refine} stage, exploiting the differentiability of 3DGS, we optimize the camera pose by backpropagating layout and orientation losses directly to the pose.

Extensive experiments show that CapFrame effectively grounds compositional text instructions to camera poses in diverse 3D Gaussian scenes. Across 38 real-world scenes and 135 curated instructions, it consistently produces viewpoints whose rendered frames better satisfy compositional requirements than heuristic viewpoint search baselines and adapted trajectory generation methods. Quantitative evaluation with text-image similarity metrics, MLLM judges, and a user study further confirms stronger alignment between rendered frames and textual descriptions.
\section{Related Work}
\noindent\textbf{3D Gaussian Representation.}
3D scene representations range from implicit radiance fields~\cite{park2019deepsdf, shi2020learning, mildenhall2021nerf} to explicit structures such as meshes~\cite{kato2018neural, kanazawa2018learning}, point clouds~\cite{qi2017pointnet++, li2020end}, and voxels~\cite{schwarz2022voxgraf, sun2022direct}. Addressing the rendering latency of implicit models and the topological rigidity of explicit ones, 3D Gaussian Splatting (3DGS)~\cite{kerbl20233d} represents scenes as anisotropic Gaussian primitives. With real-time rasterization and high visual fidelity, 3DGS has become a popular representation for novel view synthesis. Subsequent work applies 3DGS to tasks including 3D segmentation~\cite{ye2024gaussian, jain2024gaussiancut, choi2024click}, scene editing~\cite{chen2024gaussianeditor, wu2024gaussctrl, wen2025intergsedit}, generative content creation~\cite{yi2024gaussiandreamer, zhou2024dreamscene360, chen2024text}, and dynamic scene modeling~\cite{luiten2024dynamic, wu20244d}. Systems such as MonoGS~\cite{matsuki2024gaussian} further show that differentiable rasterization in 3DGS enables robust 6-DoF pose optimization. Building on this property, we exploit differentiable 3DGS rendering to optimize camera viewpoints aligned to text instruction.
\\\\
\noindent \textbf{Vision-Language Understanding.}
Vision-Language Models (VLMs) learn joint visual-textual embeddings from image-text pairs. CLIP~\cite{radford2021learning} established strong contrastive alignment, followed by models improving multimodal understanding~\cite{li2022blip, li2023blip, zhai2023sigmoid, tschannen2025siglip, xie2025fg, dai2023instructblip}. To extend such semantics to 3D scenes, recent work~\cite{qin2024langsplat, wu2024opengaussian, shi2024language, li2025langsplatv2, liu2025reasongrounder} distills language features into 3D Gaussian primitives, enabling open-vocabulary querying and amodal reasoning under occlusion. However, these approaches remain largely object-centric, focusing on \textit{what} to attend to rather than \textit{how} to frame it. Meanwhile, Multimodal Large Language Models (MLLMs)~\cite{bai2025qwen3, grattafiori2024llama, liu2024llavanext, yang2023dawn} exhibit strong zero-shot spatial reasoning, enabling tasks like object reasoning~\cite{yang2025opengs} and physical simulation~\cite{zhao2025physsplat}. We study viewpoint alignment conditioned on compositional language, using MLLMs to convert photographic instructions into geometric pseudo supervisions for camera viewpoint optimization.
\\\\
\noindent \textbf{Camera Control.}
Camera control in virtual environments aims to satisfy cinematic and narrative constraints. Early methods relied on mathematical formulations~\cite{blinn2002looking, lino2015intuitive, galvane2015camera} or rule-based systems encoding cinematic heuristics~\cite{drucker1992cinema, galvane2014narrative}. Recent learning-based approaches~\cite{courant2024exceptional, li2024director3d, liu2024chatcam, zhang2025gendop} generate language-conditioned camera motion from large-scale data. For example, ChatCam~\cite{liu2024chatcam} enables conversational camera navigation, while GenDoP~\cite{zhang2025gendop} synthesizes cinematic trajectories. Optimization-based methods such as JAWS~\cite{wang2023jaws} and SplaTraj~\cite{liu2024splatraj} refine camera paths directly in 3D representations via visibility objectives. In particular, SplaTraj~\cite{liu2024splatraj} combines 3DGS with continuous language fields~\cite{qin2024langsplat} to maintain visibility of open-vocabulary targets. However, these methods do not address fine-grained compositional constraints for intentional photographic framing. 

\section{Preliminaries: 3D Gaussian Splatting}
\label{sec:preliminaries}

3D Gaussian Splatting (3DGS)~\cite{kerbl20233d} represents a scene as anisotropic Gaussians $\mathcal{G}$, each with mean $\mu_W\in\mathbb{R}^3$ and covariance $\Sigma_W\in\mathbb{R}^{3\times3}$:
\begin{equation}
G(x)=\exp\!\left(-\frac{1}{2}(x-\mu_W)^{\top}\Sigma_W^{-1}(x-\mu_W)\right).
\label{eq:3dgs_def}
\end{equation}
We parameterize $\Sigma_W$ as $\Sigma_W=RSS^{\top}R^{\top}$ to ensure validity.

Let $\bm{T}_{CW}\in SE(3)$ be the 6-DoF camera pose. During rendering, each Gaussian is projected to the image plane with
\begin{equation}
\mu_I = \pi(\bm{T}_{CW}\mu_W), \qquad
\Sigma_I = J \bm{R} \Sigma_W \bm{R}^{\top} J^{\top},
\label{eq:ewa_project}
\end{equation}
where $\pi(\cdot)$ is perspective projection, $\bm{R}$ is rotation of $\bm{T}_{CW}$, and $J$ is projection Jacobian. Pixel color is obtained by alpha compositing $\mathcal{N}$ overlapping Gaussians:
\begin{equation}
C = \sum_{k\in\mathcal{N}} c_k \alpha_k \prod_{j=1}^{k-1}(1-\alpha_j),
\label{eq:alpha_blend}
\end{equation}
with $c_k$ and $\alpha_k$ the color and opacity of the $k$-th Gaussian.

Since rasterization is differentiable, gradients propagate to $\bm{T}_{CW}$. Following MonoGS~\cite{matsuki2024gaussian}, for a pose-dependent function $f$, we define the manifold derivative using $\tau\in\mathfrak{se}(3)$:
\begin{equation}
\frac{D f(\bm{T}_{CW})}{D \bm{T}_{CW}}
=
\lim_{\tau\to0}
\frac{
\log\!\left(
f(\exp(\tau)\circ\bm{T}_{CW})
\circ
f(\bm{T}_{CW})^{-1}
\right)
}{\tau}.
\label{eq:manifold_deriv}
\end{equation}
This enables gradient-based optimization of 6-DoF camera poses.

\section{Text-Instructed Viewpoint Grounding in 3DGS}
\label{sec:task}

We formalize the task of \textit{Text-Instructed Viewpoint Grounding} (TIVG) within a 3D Gaussian scene. Let $\mathcal{S} = \{G_i\}_{i=1}^{N}$ denote a 3DGS scene reconstructed from a set of training images $\mathcal{C}$, and let $T$ be a natural language instruction describing a desired viewpoint. A camera pose, comprising a rotation matrix $\bm{R} \in SO(3)$ and a translation vector $\bm{t} \in \mathbb{R}^3$, determines the differentiable rasterization $\mathcal{R}$ of the scene:
\begin{equation}
\bm{I} = \mathcal{R}(\mathcal{S}, \bm{T}_{CW}), \quad \text{where} \quad 
\bm{T}_{CW} =
\begin{bmatrix}
\bm{R} & \bm{t} \\
\mathbf{0}^\top & 1
\end{bmatrix} \in SE(3).
\end{equation}

The objective is to identify a camera pose $\bm{T}_{CW}$ that produces a rendered image $\bm{I} \in \mathbb{R}^{3\times H\times W}$ aligned with the instruction $T$. For instance, as illustrated in \cref{fig:task}, given the instruction ``\textit{A large brown teddy bear sits on the right side, showing left view to the camera}'', the system should not only localize the bear but also determine a precise 6-DoF configuration that satisfies both the orientation (left view) and the layout (right side of the frame). 

Unlike object-centric localization~\cite{qin2024langsplat, wu2024opengaussian}, TIVG does not generally admit a unique solution, as multiple viewpoints may satisfy the same compositional description. Given this inherent non-uniqueness, we evaluate this task using text-image similarity scores computed by external VLMs, alignment ratings from two MLLM judges, together with perceptual user studies, as detailed in~\cref{sec:experiments}.

\section{Method}
\label{sec:method}
\begin{figure}[tb]
  \centering
  \includegraphics[width=\linewidth]{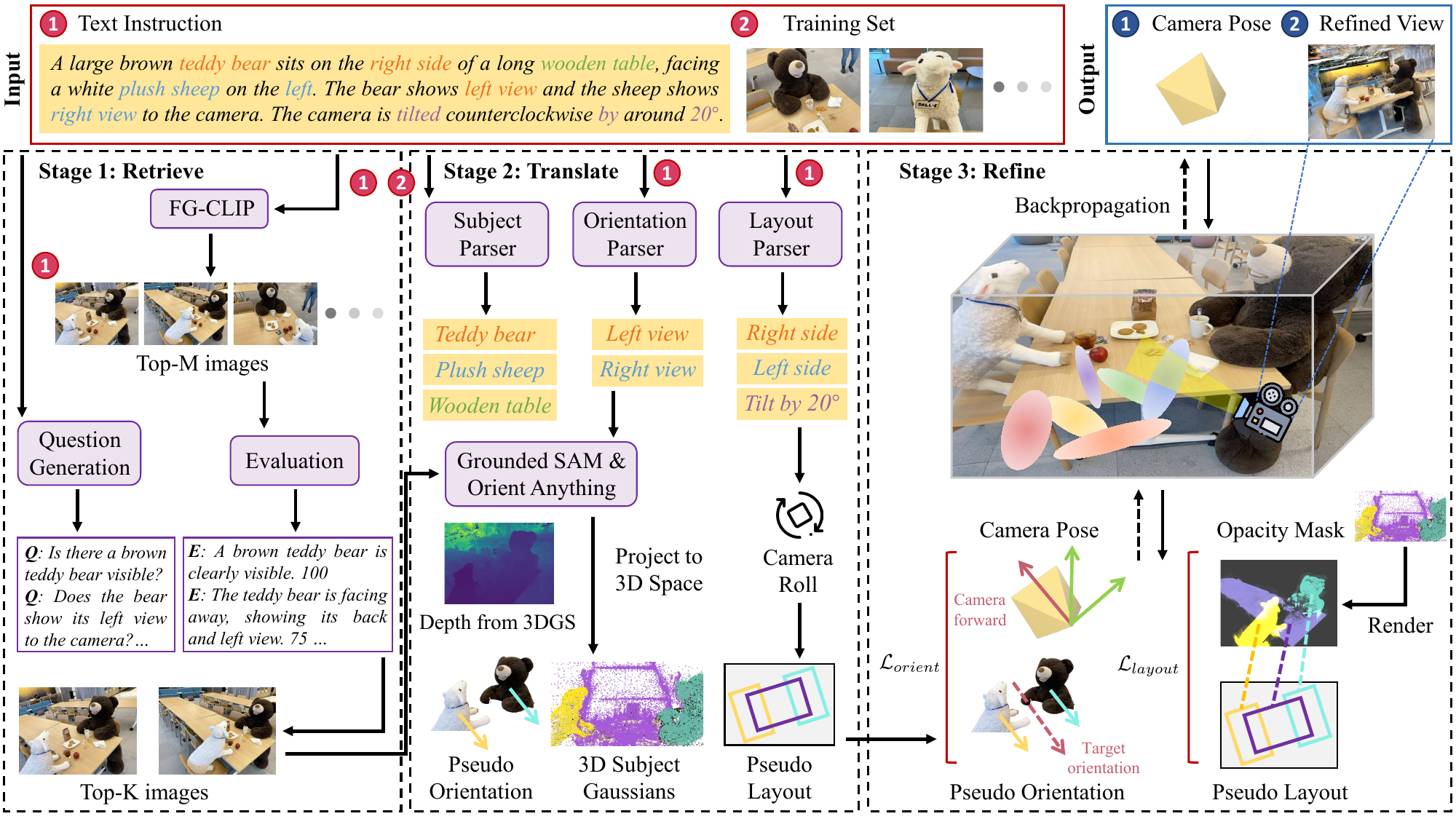}
  \caption{\textbf{Overview of CapFrame.} Starting from compositional text and views in training set, we retrieve images consistent with the text. A Question-Evaluation process within the MLLM is introduced to perform fine-grained ranking, selecting relevant image subset. Subsequently, the MLLM translates the text into geometric regularizers. In conjunction with pretrained models, we construct geometric pseudo labels based on the image subset for orientation and layout. Finally, guided by these pseudo labels, we refine camera poses through differentiable optimization in the 3D Gaussian scene.}
  \label{fig:framework}
\end{figure}

We introduce \textbf{CapFrame} to bridge the gap between abstract instructions and precise 6-DoF poses via a three-stage pipeline (\cref{fig:framework}):
\textbf{(1) Retrieve:} We identify and rank semantic anchor views from training images $\mathcal{C}$ via a \textit{Question-Evaluation} (QE) process using an MLLM (\cref{sec:retrieval}).
\textbf{(2) Translate:} We convert linguistic constraints into geometric pseudo labels for orientation and layout (\cref{sec:translate}).
\textbf{(3) Refine:} Initialized by retrieved poses, we optimize the camera by backpropagating layout and orientation losses to pose parameters (\cref{sec:refine}).

\subsection{Retrieve: Semantic-aware Viewpoint Initialization}
\label{sec:retrieval}

Exhaustively searching the continuous $SE(3)$ space of a 3DGS scene is computationally prohibitive. Moreover, gradient-based camera pose optimization requires a suitable initialization to ensure stable convergence. Since 3DGS scenes are reconstructed from a finite set of training images $\mathcal{C}$ that provide near-complete scene coverage, we assume that these existing views form a suitable discrete subspace for initializing the optimization. Therefore, in this \textbf{Retrieve} stage, we aim to identify the image from $\mathcal{C}$ most relevant to the input text instruction.

Given a text instruction $T$, we first perform global semantic alignment using FG-CLIP~\cite{xie2025fg} to identify a relevant subset of candidate poses. Since reasoning with MLLMs later is computationally expensive, this lightweight filtering step significantly reduces the candidate space. We extract a text embedding $f_T$ and image embeddings $f_{I,i}$ for each view $I_i \in \mathcal{C}$. Similarity is computed via cosine similarity $\langle f_T, f_{I,i} \rangle$, and the top-$M$ view-pose pairs are retained as
\begin{equation}
\mathcal{C}_g = \{ (I_i, \bm{T}_{CW,i}) \}_{i=1}^{M},
\end{equation}
where $M \approx |\mathcal{C}|/8$. While efficient, our preliminary experiments reveal that simple VLM-based global alignment often fails in cluttered scenes with multiple objects, as it struggles to discriminate nuanced compositional requirements.

To overcome the limitations of holistic VLM embeddings, we introduce a \textbf{Question-Evaluation (QE)} strategy that leverages the reasoning of multimodal large language models (MLLMs), such as Qwen3-VL~\cite{bai2025qwen3}. Instead of relying on a single similarity score, QE decomposes $T$ into a set of questions $\mathcal{Q}$ targeting specific semantic and photographic attributes (\eg, ``\textit{Is the subject visible?}'' or ``\textit{Is the subject located on the right side of the frame?}''). These questions are automatically generated by MLLM using a fixed, task-agnostic prompt template, ensuring a consistent and hands-free evaluation process across diverse scenes. 

The MLLM then evaluates each view $I_i \in \mathcal{C}_g$ against $\mathcal{Q}$ to produce granular scores. By averaging these scores, we obtain a ranking that reflects complex compositional alignment. The top-$K$ view-pose pairs are retained as
\begin{equation}
\mathcal{C}_f = \{ (I_i, \bm{T}_{CW,i}) \}_{i=1}^{K}.
\end{equation}
This reasoning-based filtering provides a high-quality initialization for the subsequent optimization stage, ensuring that the starting viewpoint already satisfies basic semantic constraints.
\subsection{Translate: From Language to Geometric Pseudo Labels}
\label{sec:translate}

The \textbf{Retrieve} stage provides a strong initialization by selecting candidate viewpoints from the training set. Our goal, however, is to search the \emph{continuous} camera pose space in $SE(3)$ and find a viewpoint whose rendered frame matches the text instruction. Because natural language rarely specifies optimization-ready numeric targets (\eg, exact angles or pixel coordinates), directly constructing differentiable objectives from text is challenging. We therefore introduce a translation step that converts compositional instructions into \emph{geometric pseudo labels}—structured targets guiding continuous pose refinement in 3DGS.

We follow key decisions in photographic composition: selecting salient subjects, defining viewing direction, and arranging subjects within the frame. Accordingly, we extract subject Gaussians and construct two pseudo-label types: (i) orientation pseudo labels for viewpoint control and (ii) layout pseudo labels for framing constraints.
\\\\
\noindent\textbf{Subject--Gaussian Association.}
\label{subsec:keysub}
Let $\mathcal{C}_f=\{(I_j,\bm{T}_{CW,j})\}_{j=1}^{K}$ denote the top-$K$ retrieved view-pose pairs. Given instruction $T$, we prompt an MLLM to extract key subjects
$\mathcal{O}=\{O_1,\dots,O_n\}$, including mentioned entities and visually dominant objects affecting framing. Each subject is associated with a subset of 3D Gaussians to enable subject-centric reasoning and differentiable mask construction. Unlike open-vocabulary localization methods that embed language features into all Gaussians~\cite{qin2024langsplat, wu2024opengaussian}, we localize only the small set $\mathcal{O}$, which suffices for constructing pseudo labels.

For each subject $O_i$ and retrieved view $I_j$, we obtain a segmentation mask $M^{\text{SAM}}_{ij}$ using Grounded SAM~\cite{kirillov2023segment, liu2024grounding, ren2024grounded}. Pixels $(u,v)\in M^{\text{SAM}}_{ij}$ are back-projected using depth rendered from 3DGS to obtain a 3D point set
\begin{equation}
\mathcal{P}_{ij}=\{(X_W,Y_W,Z_W)^\top \mid (u,v)\in M^{\text{SAM}}_{ij}\}.
\end{equation}
Since rendered depth may be noisy, these points may not coincide exactly with Gaussian means. We therefore retrieve a subject-specific Gaussian subset $\mathcal{G}_{Q,i}\subset\mathcal{G}$ via KNN search between $\cup_j\mathcal{P}_{ij}$ and Gaussian means $\{\mu_{W,k}\}$. The subject centroid is computed as
\begin{equation}
\bm{C}_{W,i}=\frac{1}{|\mathcal{G}_{Q,i}|}\sum_{k\in\mathcal{G}_{Q,i}}\mu_{W,k}.
\end{equation}
Rendering only $\mathcal{G}_{Q,i}$ yields a differentiable subject mask for layout objectives.
\\\\
\noindent\textbf{Orientation Pseudo Labels.}
\label{sec:orientation}
Many instructions specify viewpoint and subject orientation through cues such as \textit{front view}, \textit{side view}, or \textit{look down}. We convert such language into orientation pseudo labels by combining (i) relative offsets inferred from {\it text} and (ii) estimated subject orientation from images.

Given $T$ and $\mathcal{O}$, we prompt the MLLM to infer relative orientation offsets
$\Delta\bm{d}_i=(\Delta\phi_i,\Delta\theta_i,\Delta\gamma_i)$,
representing azimuth, elevation and roll adjustments. Orientation constraints are applied only to asymmetric subjects with meaningful front or back semantics. Symmetric objects (\eg, balls) rely solely on layout constraints (subject types are identified by MLLM).

To estimate current orientation of {\it subjects}, we apply Orient Anything~\cite{wang2024orient} to the top-$K$ images, obtaining
$(\phi_{O,i},\theta_{O,i})$ ($\gamma_{O,i}$ is not used as we only need the forward direction of the subject).
The target orientation combines this estimate with the text-derived offsets:
\begin{equation}
\bm{P}^{\text{orient}}_{O,i}=
(\phi_{O,i}+\Delta\phi_i,\ \theta_{O,i}+\Delta\theta_i,\ \Delta\gamma_i),
\label{eq:ori_O}
\end{equation}
where azimuth is wrapped modulo and elevation clamped to a valid range. The result is {\it mapped to world coordinates}, producing $\bm{P}^{\text{orient}}_{W,i}$ used as a differentiable regularizer during refinement.

\subsubsection{Layout Pseudo Labels.}
\label{sec:layout}

Photographic intent also specifies subject placement within the frame (\eg, \textit{subject on the right}, \textit{centered}, \textit{close-up}). We therefore prompt the MLLM to infer a target 2D layout in normalized image coordinates. For each subject $O_i$, the layout pseudo label is a bounding box
\begin{equation}
\bm{P}^{\text{layout}}_i=[x_{\min},y_{\min},x_{\max},y_{\max}],
\label{eq:com}
\end{equation}
which is converted to pixel coordinates using image width $W$ and height $H$.

To remain consistent with roll constraints, we adjust the layout if $\Delta\gamma_i\neq 0$. Let $\bm{c}=(c_x,c_y)$ denote the center of the bounding box and $\bm{b}$ denote a corner of the bounding box.
Each corner is rotated around $\bm{c}$ by $\Delta\gamma_i$:
\begin{equation}
\bm{b}_{\gamma}=\bm{c}+\bm{R}(\Delta\gamma_i)(\bm{b}-\bm{c}),
\end{equation}
where $\bm{R}(\Delta\gamma_i)$ is the 2D rotation matrix. The rotated corners define the final layout target used during refinement.

\subsection{Refine: Gradient-Based Camera Pose Optimization}
\label{sec:refine}
Starting from the pose retrieved in~\cref{sec:retrieval}, we optimize an incremental update $\tau=[\Delta\bm{r},\Delta\bm{t}]^\top\in\mathfrak{se}(3)$ and update the camera pose by left composition
$\bm{T}_{CW}\leftarrow \exp(\tau)\circ \bm{T}_{CW}$.
We minimize a multi-objective loss
\begin{equation}
\mathcal{L} = \mathcal{L}_{\text{layout}} + \mathcal{L}_{\text{orient}},
\end{equation}
where both terms are derived from pseudo labels in~\cref{sec:translate}.
\\\\
\noindent\textbf{Layout Loss.}
\label{sec:layoutloss}
The layout loss enforces composition constraints by encouraging each subject to occupy its target region $\bm{P}^{\text{layout}}_i$ (\cref{sec:layout}). For subject $O_i$, we render only its Gaussian subset $\mathcal{G}_{Q,i}$ and obtain a differentiable opacity mask $M_i^g\in[0,1]^{H\times W}$ via alpha compositing:
\begin{equation}
M_i^g(u,v) = \sum_{k=1}^{N_i} \alpha_k(u,v)\prod_{j<k}\big(1-\alpha_j(u,v)\big),
\end{equation}
where $N_i$ is the number of contributing Gaussians and $\alpha_k(u,v)$ denotes the opacity of the $k$-th Gaussian. Let $\bm{C}_{P,i}$ denote the soft centroid of $M_i^g$:
\begin{equation}
\bm{C}_{P,i} =
\frac{\sum_{(u,v)\in\Omega} (u,v)\, M_i^g(u,v)}
{\sum_{(u,v)\in\Omega} M_i^g(u,v) + \epsilon},
\end{equation}
where $\Omega$ is the image domain. If $\bm{C}_{P,i}$ lies inside $\bm{P}^{\text{layout}}_i$, the loss is zero; otherwise we penalize the distance to the nearest point on the region boundary $\partial P^{\text{layout}}_i$:
\begin{equation}
\mathcal{L}_{\text{center}}^{i} =
\begin{cases}
0, & \text{if } \bm{C}_{P,i} \in \bm{P}^{\text{layout}}_i, \\
\min_{\bm{p} \in \partial P^{\text{layout}}_i}
\left\|
\bm{C}_{P,i}-\bm{p}
\right\|_2, & \text{otherwise}.
\end{cases}
\end{equation}

Empirically, centroid constraints alone are insufficient when subjects are elongated or partially outside the target region. We therefore additionally encourage mask coverage inside the box and penalize leakage outside it. Using the indicator $\mathbb{I}(\cdot)$, we define
\begin{equation}
\begin{aligned}
&\mathcal{L}^{i}_{\text{in}} =
1-\frac{
\sum_{(u,v)\in\Omega} M_i^g(u,v)\,
\mathbb{I}\!\big((u,v)\in \bm{P}^{\text{layout}}_i\big)
}{
\sum_{(u,v)\in\Omega} M_i^g(u,v) + \epsilon
},\\
&\mathcal{L}^{i}_{\text{out}} =
\frac{
\sum_{(u,v)\in\Omega} M_i^g(u,v)\,
\mathbb{I}\!\big((u,v)\notin \bm{P}^{\text{layout}}_i\big)
}{
\sum_{(u,v)\in\Omega} M_i^g(u,v) + \epsilon
}.
\end{aligned}
\end{equation}

The total layout loss aggregates all subjects:
\begin{equation}
\mathcal{L}_{\text{layout}}=
\sum_{i=1}^{n}
\left(
\lambda_c\mathcal{L}^{i}_{\text{center}}
+\lambda_{\text{in}}\mathcal{L}^{i}_{\text{in}}
+\lambda_{\text{out}}\mathcal{L}^{i}_{\text{out}}
\right),
\end{equation}
where $\lambda_c$, $\lambda_{\text{in}}$, and $\lambda_{\text{out}}$ are scalar weights and $\epsilon$ ensures numerical stability.
\\\\
\noindent\textbf{Orientation Loss.}
\label{sec:orientloss}
The orientation loss enforces viewpoint constraints from orientation pseudo labels (\cref{sec:orientation}). It contains three terms: gravity (discouraging tilt), forward-facing (aligning the camera with the desired viewpoint), and look-at (keeping the subject visible).

\paragraph{Gravity.}

When roll is not specified (\ie, $\Delta\gamma_i=0$), we encourage an upright camera by aligning the camera up direction $\bm{D}_{C,cu}$ with the world up direction in the camera frame $\bm{D}_{C,wu}$:
\begin{equation}
\mathcal{L}_{\text{gravity}} = 1 - \langle \bm{D}_{C,cu}, \bm{D}_{C,wu} \rangle .
\end{equation}
If roll is specified, this term is disabled and roll is handled by the roll-corrected layout labels (\cref{sec:layout}).

\paragraph{Forward-facing.}

For asymmetric subjects $O_i\in\mathcal{O}_w$, we align the camera forward direction $\bm{D}_{C,cf}$ with the target orientation $\bm{P}^{\text{orient}}_{C,i}$:
\begin{equation}
\mathcal{L}^{i}_{\text{forward}} =
1 - \langle \bm{D}_{C,cf}, \bm{P}^{\text{orient}}_{C,i} \rangle .
\end{equation}

\paragraph{Look-at.}

To keep the subject in view, we align the camera with the subject centroid. Let $\bm{C}_{cam}$ and $\bm{C}_{W,i}$ denote the camera center and subject centroid (\cref{subsec:keysub}). The normalized direction
\begin{equation}
\bm{d}_{W,i} =
\frac{\bm{C}_{W,i}-\bm{C}_{cam}}{\|\bm{C}_{W,i}-\bm{C}_{cam}\|_2}
\end{equation}
is expressed in the camera frame as $\bm{D}_{C,cam,i}$, yielding
\begin{equation}
\mathcal{L}^{i}_{\text{look-at}} =
1 - \langle \bm{D}_{C,cf}, \bm{D}_{C,cam,i} \rangle .
\end{equation}

The final orientation loss is
\begin{equation}
\mathcal{L}_{\text{orient}}
=
\lambda_g \mathcal{L}_{\text{gravity}}
+
\sum_{O_i \in \mathcal{O}_w}
\left(
\lambda_f \mathcal{L}^{i}_{\text{forward}}
+
\lambda_l \mathcal{L}^{i}_{\text{look-at}}
\right),
\end{equation}
where $\lambda_g$, $\lambda_f$, and $\lambda_l$ are scalar weights.

\section{Results}
\label{sec:experiments}
\begin{figure}[htbp]
  \centering
  \includegraphics[width=\linewidth]{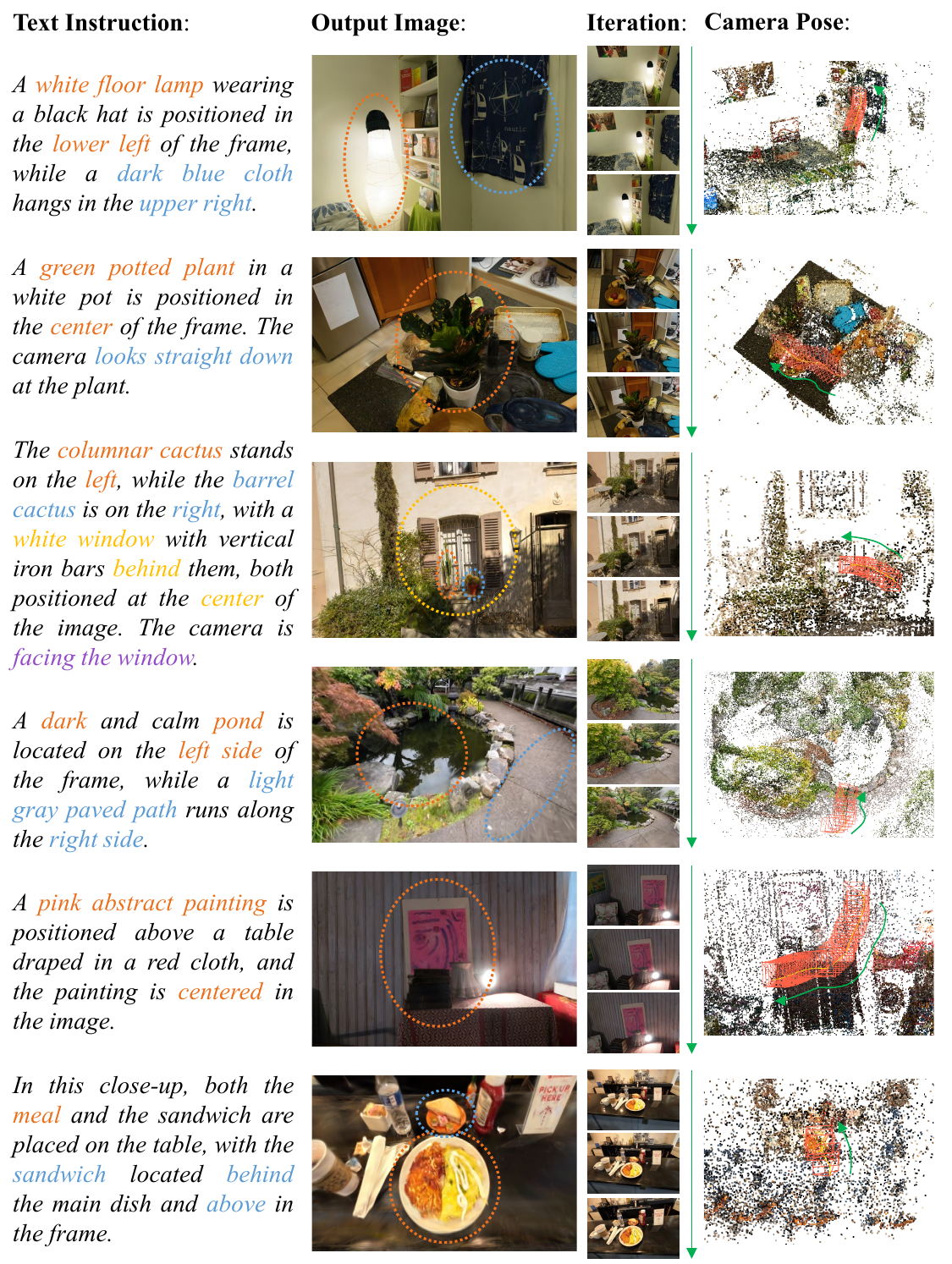}
  \caption{\textbf{Text-instructed viewpoint grounding in diverse 3DGS scenes.}
Given a natural language instruction (left), CapFrame identifies a camera pose that produces a frame consistent with the described subject, composition and viewpoint.
Starting from retrieved initial views, the camera pose is refined through differentiable optimization in the 3D Gaussian scene to generate the final rendered frame (middle), with the optimized camera pose shown on the right.}
  \label{fig:capture}
\end{figure}
\begin{figure}[tb]
  \centering
  \includegraphics[width=\linewidth]{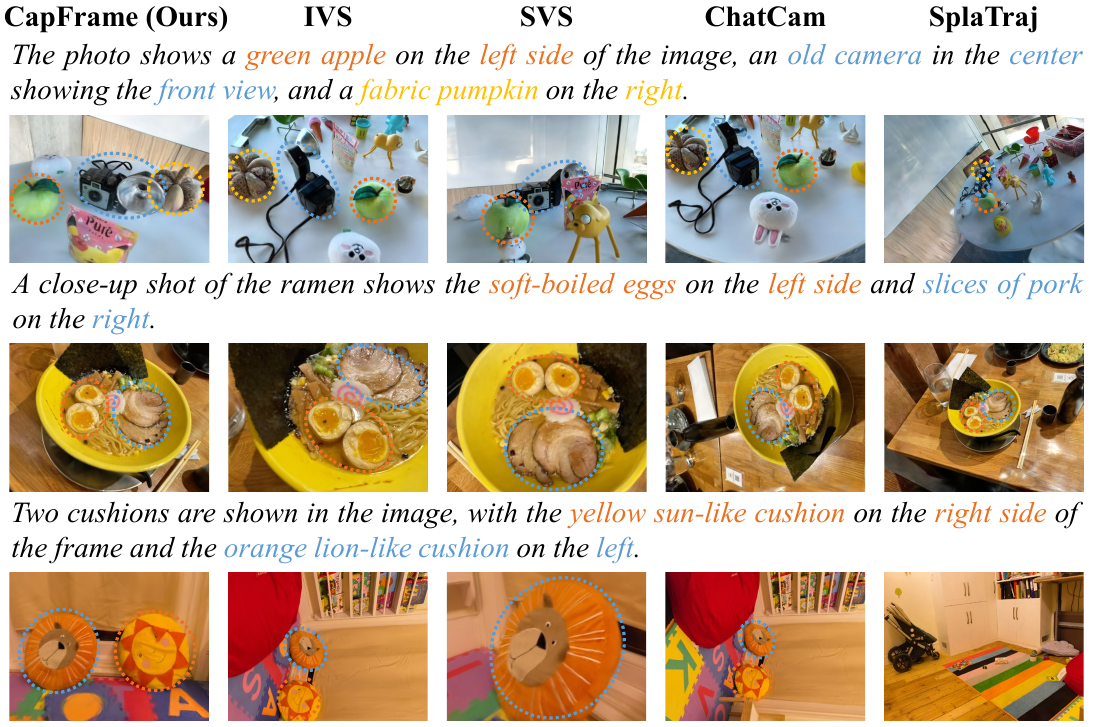}
  \caption{\textbf{Qualitative comparisons.} CapFrame renders images aligned with specified subjects, composition and viewpoint, whereas other baselines are largely limited to object localization and often fail to satisfy composition and viewpoint requirements.}
  \label{fig:qualitative}
\end{figure}
\noindent\textbf{Implementation Details.}
We use Qwen3-VL~\cite{bai2025qwen3} as the MLLM for text reasoning in the \textbf{Retrieve} stage and pseudo-label generation in the \textbf{Translate} stage. 
For scene representation, we adopt the standard 3DGS implementation~\cite{kerbl20233d} with camera pose gradients from MonoGS~\cite{matsuki2024gaussian} (\cref{eq:manifold_deriv}). As our focus is viewpoint grounding, the reconstruction method is not critical. In \textbf{Retrieve}, the number of top-ranked views for fine-grained ranking is set to $K=2$. Camera pose refinement in \textbf{Refine} is optimized using Adam~\cite{adam2015} with learning rates $0.007$ (rotation) and $0.005$ (translation). We set $\lambda_c=1$ and $\lambda_{\text{in}}=6$, while all other weights are $2$ ($\lambda_{\text{out}},\lambda_g,\lambda_f,\lambda_l$). Optimization runs for up to $1500$ iterations with early stopping when the loss change falls below $1\times10^{-4}$ or $10^{-5}$ depending on scene scale. Experiments run on a single NVIDIA H100 GPU (80GB).
\\\\\noindent\textbf{Baselines.}
As no established baselines exist for text-instructed viewpoint grounding in 3DGS scenes, we construct two heuristic baselines inspired by scene exploration methods~\cite{skartados2024finding}: \textit{Interpolation-based Viewpoint Search} (IVS) and \textit{Sampling-based Viewpoint Search} (SVS). IVS generates candidate viewpoints by interpolating between top-$K$ retrieved camera poses from the \textbf{Retrieve} stage (\cref{sec:retrieval}). SVS instead samples camera poses densely on a sphere centered around the key subjects identified in the \textbf{Translate} stage (\cref{sec:translate}). For both baselines, each candidate pose is rendered using 3DGS, and the final viewpoint is selected as the frame with the highest text-image similarity measured by FG-CLIP~\cite{xie2025fg}. In addition, we adapt the relevant components of two trajectory generation methods, ChatCam~\cite{liu2024chatcam} and SplaTraj~\cite{liu2024splatraj}, to our TIVG setting. We build on ChatCam's available \textit{Anchor Determination}, which optimizes camera poses via CLIP~\cite{radford2021learning} gradients, and reproduce the relevant part of SplaTraj.
\\\\\noindent\textbf{Datasets.}
We evaluate on real-world 3D reconstruction datasets: 5 scenes from {\it Mip-NeRF 360}~\cite{barron2022mip}, 13 from {\it Deep Blending}~\cite{hedman2018deep}, 4 from {\it Tanks and Temples}~\cite{knapitsch2017tanks}, 4 from {\it LERF-OVS}~\cite{kerr2023lerf}, and 12 from {\it DL3DV-10K}~\cite{ling2024dl3dv}, totaling 38 indoor and outdoor scenes. For scenes without camera poses, we estimate them by COLMAP~\cite{schonberger2016structure}. We manually curate 3-4 instructions per scene, resulting in 135 text descriptions.
\\\\\noindent\textbf{Metrics.}
As text-instructed viewpoint grounding lacks a unique ground-truth viewpoint, evaluation is non-trivial. For quantitative evaluation, we compute text-image similarity using CLIP~\cite{radford2021learning} (ViT-H/14) and SigLIP2~\cite{tschannen2025siglip}. Using two VLMs reduces bias from a single scoring function and provides a more robust estimate of semantic alignment. Similarly, we introduce two MLLM judges (GPT-5.4-mini~\cite{gpt2026openai} and Gemini-2.5-Flash~\cite{comanici2025gemini}) as blind photographic evaluators that rate the compositional \textit{alignment scores} (AS) and select the most aligned output frame across methods, computing the \textit{win rates} (WR) of CapFrame over the baselines. We additionally conduct a user study to evaluate perceptual alignment. The study involves 33 participants, each completing questionnaires with 12 groups. In each group, participants rate three candidate frames on a 5-point scale and select the frame that best matches the instruction.\\
\begin{table}[t]
\caption{\textbf{Quantitative comparisons.} The left two columns report text-image similarity scores from two VLMs (CLIP and SigLIP2). The middle four columns report alignment scores (AS) and win rates (WR) from two MLLM judges (GPT and Gemini). The right two columns summarize user study results, including average rating and preference percentage. ``\textemdash'' indicates that the corresponding baselines are not evaluated in our user study.}
\label{tab:quantitative}

\centering
\small
\setlength{\tabcolsep}{4.5pt}
\renewcommand{\arraystretch}{1.15}

\resizebox{\linewidth}{!}{
\begin{tabular}{l|cc|cccc|cc}
\toprule
\multirow{2}{*}{Method} 
& \multicolumn{2}{c|}{VLM metrics} 
& \multicolumn{4}{c|}{MLLM judges} 
& \multicolumn{2}{c}{User study} \\
\cmidrule(lr){2-3}\cmidrule(lr){4-5}\cmidrule(lr){6-7}\cmidrule(lr){8-9}
& CLIP$\uparrow$ 
& SigLIP2$\uparrow$ 
& GPT AS$\uparrow$ 
& GPT WR$\uparrow$ 
& Gemini AS$\uparrow$ 
& Gemini WR$\uparrow$ 
& Rating$\uparrow$ 
& Preference$\uparrow$ \\
\midrule
SplaTraj~\cite{liu2024splatraj} 
& 0.241 & 0.276 & 3.08 & 3.0\% & 2.78 & 3.0\% & \textemdash & \textemdash \\

ChatCam~\cite{liu2024chatcam} 
& 0.258 & 0.283 & 3.42 & 3.7\% & 3.27 & 5.2\% & \textemdash & \textemdash \\

SVS 
& 0.275 & 0.435 & 4.06 & 14.8\% & 3.88 & 16.3\% & 2.52 & 16.7\% \\

IVS 
& 0.265 & 0.396 & 3.49 & 5.9\% & 3.19 & 8.1\% & 2.53 & 5.6\% \\

\textbf{CapFrame (Ours)} 
& \textbf{0.282} & \textbf{0.448} 
& \textbf{4.75} & \textbf{72.6\%} 
& \textbf{4.22} & \textbf{67.4\%} 
& \textbf{4.50} & \textbf{77.7\%} \\
\bottomrule
\end{tabular}
}
\end{table}

\noindent\textbf{Text-Instructed Viewpoint Grounding.}
As shown in~\cref{fig:capture}, CapFrame grounds compositional text instructions to camera poses, enabling automatic viewpoint selection in diverse 3D Gaussian scenes. By leveraging the zero-shot reasoning ability of MLLMs, the framework generalizes to varied scenarios and supports intuitive text-guided exploration of complex 3D environments.
\\\\\noindent\textbf{Comparisons with Baselines.}
\Cref{fig:qualitative} presents qualitative comparisons under identical instructions. While IVS and SVS often retrieve viewpoints where the target subject is visible, they frequently fail to satisfy the required composition or orientation. For example, IVS places the \textit{green apple} away from the left-side position, while SVS fails to place the \textit{slices of pork} on the right. IVS can also inherit biases from retrieved camera poses: if the original views are tilted, interpolated viewpoints may preserve such unnatural orientations, as observed in the \textit{two cushions} example. In the same example, ChatCam~\cite{liu2024chatcam} produces a tilted view of the \textit{cushions}, as CLIP-based~\cite{radford2021learning} guidance is invariant to orientation. SplaTraj~\cite{liu2024splatraj} settles on a viewpoint where the \textit{cushions} are no longer visible, because its learned language field~\cite{qin2024langsplat} fails to distinguish the target objects. In contrast, CapFrame optimizes camera poses using geometric pseudo labels encoding both layout and orientation, producing frames better aligned with the compositional intent of the instruction. Quantitative results in~\cref{tab:quantitative} further show that CapFrame achieves the highest text-image alignment on VLM-based similarity. The same trend holds across two MLLM judges, where CapFrame achieves the best alignment score and win rate. Moreover, our user study reveals a consistent preference for CapFrame over IVS and SVS.
\\\\
\noindent\textbf{Component Analysis.}
\begin{table}[t]
\centering
\caption{\textbf{SigLIP2 score} before and after camera pose refinement.}
\setlength{\tabcolsep}{4.2pt}
\begin{tabular}{lccccc}
\toprule
 Stage & Mip-NeRF & Deep Blending & Tanks & LERF-OVS & DL3DV-10K \\
\midrule
Retrieve          &0.121  &0.274  &0.307 &0.503 &0.393  \\
Refine       &\textbf{0.244}  &\textbf{0.406}  &\textbf{0.339} &\textbf{0.708} &\textbf{0.500}\\
\bottomrule
\label{tab:refine}
\end{tabular}
\end{table}
We conduct studies to analyze the key components of CapFrame following the pipeline order: the Question-Evaluation (QE) strategy in \textbf{Retrieve}, the layout and orientation losses derived from geometric pseudo labels in \textbf{Translate}, and the pose refinement process in \textbf{Refine}.

\begin{figure}[tb]
  \centering
  \includegraphics[width=\linewidth]{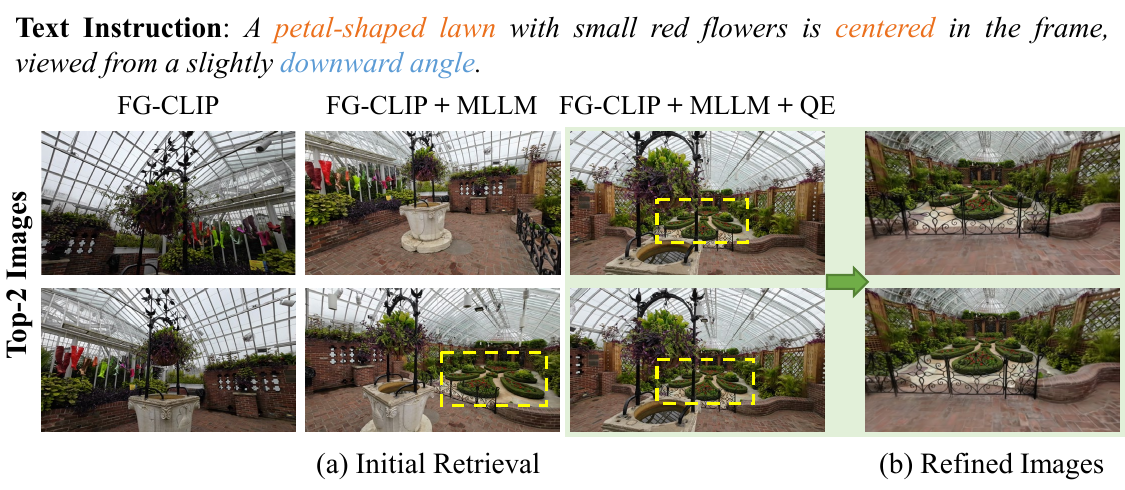}
  \caption{\textbf{Ablation of initialization.} We present results of Top-$K$ image selection across different retrieval methods. In complex scenes, (a) FG-CLIP~\cite{xie2025fg} fails to localize the subject. Without QE for multi-dimensional scoring, the subject can still be absent from the image. (b) A good initialization is essential for camera pose convergence.}
  \label{fig:initial}
\end{figure}

\begin{figure}[tb]
  \centering
  \includegraphics[width=\linewidth]{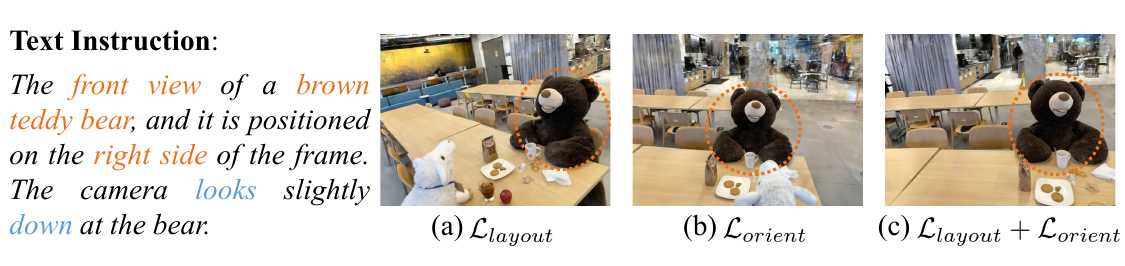}
  \caption{\textbf{Ablation of loss.} Layout and orientation losses significantly influence the optimization performance. (a) Without orientation loss, the camera fails to achieve the front view of the bear. (b) Without layout loss, the bear is not positioned on the right side. (c) With both losses, the bear satisfies the specified composition and viewpoint.}
  \label{fig:loss}
\end{figure}

We first evaluate QE in the \textbf{Retrieve} stage. As illustrated in~\cref{fig:initial}, using only FG-CLIP~\cite{xie2025fg} for coarse retrieval often returns semantically related views that do not contain the target subject, especially in cluttered scenes. Prompting the MLLM to assign a single holistic score shows similar limitations because compositional constraints are not explicitly evaluated. In contrast, QE decomposes the instruction into multiple questions, enabling the MLLM to assess candidate views along several semantic aspects. As a result, retrieved views more reliably contain the target subject and provide better initialization for pose optimization. Next, we conduct the ablation study to analyze the layout and orientation losses used in pose refinement (\cref{fig:loss}). With both losses, the optimized pose satisfies the desired composition and viewpoint. The layout loss regulates subject placement, while the orientation loss enforces the viewing direction. Removing either loss degrades alignment: without the orientation loss the camera fails to reach the specified view, and without the layout loss the subject placement deviates from the intended composition. Finally, we examine the effect of \textbf{Refine} stage. Because the desired viewpoint may not exist among retrieved views, refinement performs gradient-based camera pose optimization using geometric pseudo labels, enabling continuous pose adjustment in $SE(3)$. We measure text-image alignment using SigLIP2~\cite{tschannen2025siglip}. As shown in \cref{tab:refine}, refined views consistently achieve higher alignment scores than the retrieved initial views across all datasets.
\begin{table}[t]
\centering
\small
\setlength{\tabcolsep}{8pt}
\renewcommand{\arraystretch}{1.25}

\caption{\textbf{Sensitivity analysis under different perturbations.} We measure the deviation from the unperturbed output, where $\Delta R$ and $\Delta t$ are the rotation and translation differences and $\Delta$CLIP and $\Delta$SigLIP2 are the changes in the VLM-based metrics.}
\label{tab:sensitivity}

\resizebox{\linewidth}{!}{
\begin{tabular}{ll|cc|cc}
\toprule
Type
& Perturbation 
& $\Delta R$
& $\Delta t$
& $\Delta$CLIP
& $\Delta$SigLIP2 \\
\midrule
\multirow{1}{*}{Input text}
& Paraphrased text & 9.61$^\circ$ & 0.732 & $+0.013$ & $+0.032$ \\

\midrule
\multirow{2}{*}{Top-$K$ images}
& Top-$3$ images & 8.99$^\circ$ & 1.268 & $-0.014$ & $-0.061$ \\
& Top-$5$ images & 13.28$^\circ$ & 1.801 & $-0.018$ & $-0.165$ \\

\midrule
\multirow{3}{*}{Pseudo labels}
& Layout label removal & 4.08$^\circ$ & 5.701 & $-0.024$ & $-0.236$ \\
& Orientation label removal & 33.71$^\circ$ & 2.312 & $-0.021$ & $-0.128$ \\
& Moderate noise & 8.39$^\circ$ & 0.219 & $-0.012$ & $-0.048$ \\

\bottomrule
\end{tabular}
}
\end{table}
\\\\
\noindent\textbf{Sensitivity Analysis.} We analyze the sensitivity of CapFrame to different perturbations in \cref{tab:sensitivity}, comparing the changes in camera pose and VLM-based metrics relative to the unperturbed output. First, paraphrasing each instruction twice with GPT-5.4-mini~\cite{gpt2026openai} slightly improves the alignment scores, demonstrating insensitivity to variation in input text. Next, when initializing from lower-ranked views (Top-3 and Top-5), the pose deviation grows gradually with $K$ and the alignment scores decrease, indicating that camera refinement partially compensates for imperfect retrieval, though a good initialization remains beneficial. Finally, removing the pseudo labels reveals their complementary roles: excluding the layout label raises translation deviation, whereas excluding the orientation label increases rotation deviation. Moreover, under moderate label noise (5--10 pixels for layout and 5--10$^\circ$ for orientation), CapFrame stays close to its original output, confirming its robustness to imperfect pseudo labels.


\section{Conclusion}
We introduce a new task, text-instructed viewpoint grounding in 3D Gaussian scenes, and present CapFrame, a framework for solving it. CapFrame leverages the zero-shot reasoning of MLLMs to translate natural language instructions into geometric pseudo labels, including orientation and layout constraints, which guide camera pose refinement through differentiable optimization in 3DGS. Extensive experiments demonstrate that CapFrame generalizes across diverse scenes and produces viewpoints that more faithfully align with compositional text instructions, enabling intuitive language-driven exploration of 3D environments.
\\\\
\noindent\textbf{Limitations.} CapFrame relies heavily on MLLMs in several stages, including retrieval and translation. This design enables test-time optimization without additional training, but also introduces sensitivity to the prompts provided to the MLLM, which may affect the resulting pseudo labels and camera placement. While this work focuses on introducing the new task and establishing a first solution, future research could explore more robust prompting strategies and prompt optimization to further improve performance.

\section*{Acknowledgements}
This work was supported by DENSO IT LAB Recognition, Control, and Learning Algorithm Collaborative Research Chair (Science Tokyo) and the experiments were conducted using TSUBAME 4.0 supercomputer.

%
%
\bibliographystyle{splncs04}
\bibliography{main}

\clearpage
\appendix
\begin{center}
{\Large\bfseries Supplementary Material\par}
\vspace{0.3em}
{\large\bfseries
CapFrame: Text-Instructed Viewpoint Grounding in 3D Gaussian Scenes via Geometric Pseudo Labels
\par}
\end{center}

\vspace{1em}

In this supplementary material, we provide additional implementation details,
runtime analysis, and extended experimental results.
It is organized as follows:
\begin{itemize}
\item \cref{exp_detail}: Additional implementation details, including details of IVS and SVS and gravity-direction estimation.
\item \cref{pseudo}: Visualization of the pseudo labels produced in the \textbf{Translate} stage.
\item \cref{time}: Runtime analysis, including the runtime of the \textbf{Retrieve}, \textbf{Translate}, and \textbf{Refine} stages, as well as an acceleration strategy.
\item \cref{more_exp}: Additional results on text-instructed viewpoint grounding, including extended comparisons with baselines, component analysis, and beyond source-view initialization.
\item \cref{fail}: Failure cases and corresponding analysis.
\item \cref{prompt}: Prompts used by the MLLM in the \textbf{Retrieve} and \textbf{Translate} stages.
\item \cref{user_study}: Details of the questionnaire used in the user study.
\end{itemize}

\begin{figure}[htpb]
  \centering
  \includegraphics[width=\linewidth]{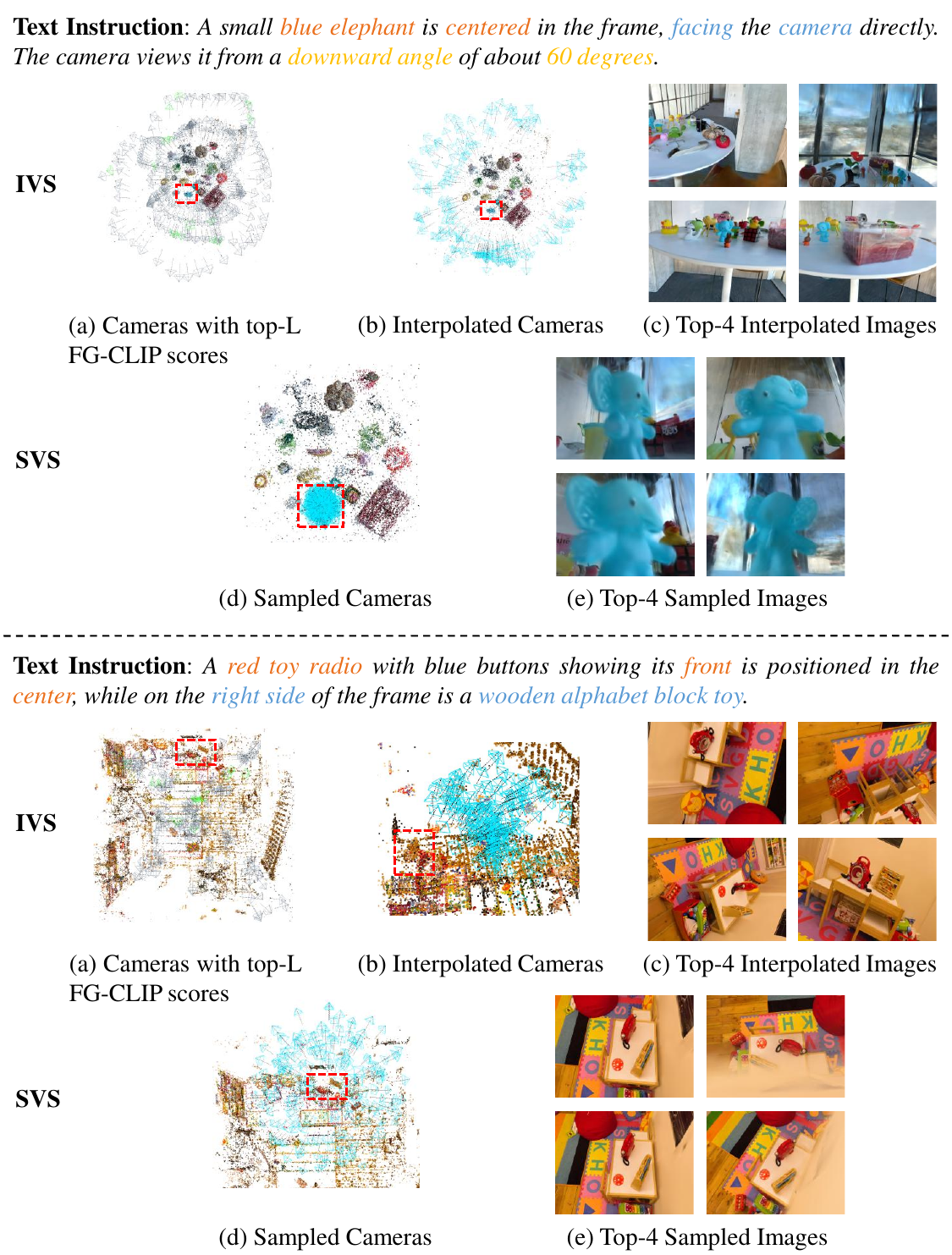}
  \caption{\textbf{Visualization of viewpoint generation in IVS and SVS.} Two examples are shown, with key subjects highlighted by red boxes. For IVS, (a) shows the top-$L$ training cameras retrieved by FG-CLIP~\cite{xie2025fg}, (b) shows the candidate cameras generated by pose interpolation, and (c) shows the top-4 rendered views ranked by FG-CLIP. For SVS, (d) shows the candidate cameras sampled around the translated subject locations, and (e) shows the top-4 rendered views ranked by FG-CLIP.}
  \label{fig:ivs_svs}
\end{figure}

\section{Implementation Details}
\label{exp_detail}
\subsection{Details of IVS and SVS}

Since there are no established baselines for \textit{text-instructed viewpoint grounding}, we design two heuristic baselines inspired by the scene exploration strategies proposed in Finding Waldo~\cite{skartados2024finding}. That work studies viewpoint search in NeRF scenes and introduces two exploration strategies: (i) interpolation between promising viewpoints (Pose Interpolation-Based Search, PIBS) and (ii) random sampling of camera poses (Guided Random Search, GRS). We adapt these two strategies to the text-conditioned setting by replacing their original scoring functions with text-image alignment and by incorporating geometric constraints derived from the reconstructed scene.

Specifically, we introduce \textit{Interpolation-based Viewpoint Search} (IVS) and \textit{Sampling-based Viewpoint Search} (SVS). IVS is derived from the interpolation strategy of PIBS, while SVS follows the sampling principle of GRS. The key difference is that our baselines are required to generate viewpoints that satisfy a \textit{text instruction}, which requires modifying how promising viewpoints are selected and how the sampling region is defined.
\\\\
\noindent\textbf{IVS (Interpolation-based Viewpoint Search).}
IVS is inspired by the pose interpolation strategy of PIBS~\cite{skartados2024finding}, which generates new candidate viewpoints by interpolating between camera poses that already achieve high task scores. In the original method, promising viewpoints are selected based on task-specific criteria such as saliency or image quality. In our text-instructed setting, we instead identify promising viewpoints according to their alignment with the text.

Concretely, similar to the \textbf{Retrieve} stage described in Sec.~5.1 of the main paper, we first select the top-$L$ training images that are most consistent with the text instruction using FG-CLIP~\cite{xie2025fg}, where $L=\min(13, N)$ and $N$ denotes the number of training images. Since viewpoints close to these highly aligned images are likely to satisfy the instruction, we interpolate between their camera poses to generate additional candidate viewpoints. Concretely, all possible pose pairs are formed among the top-$L$ views (\ie, $\frac{L(L-1)}{2}$ pairs). For each pair, three intermediate poses are generated via interpolation, resulting in $\frac{3L(L-1)}{2}$ candidate viewpoints. Rotations are interpolated in quaternion space using spherical linear interpolation (SLERP)~\cite{shoemake1985animating}, while translations are linearly interpolated. Rendering these poses produces a candidate image set.
\\\\
\noindent\textbf{SVS (Sampling-based Viewpoint Search).}
SVS is inspired by the random sampling strategy of GRS~\cite{skartados2024finding}, which explores the camera pose space by sampling viewpoints within a bounded spatial region. Unlike GRS, where sampling is guided by the distribution of training cameras, the relevant region in our task depends on the subjects mentioned in the text instruction.

To address this, we define the sampling region using the 3D Gaussian representation obtained in the \textbf{Translate} stage (Sec.~5.2 of the main paper). We first identify the Gaussians associated with key subjects mentioned in the instruction and compute their mean center. A bounding sphere enclosing these Gaussians is then constructed. Camera viewpoints are generated by sampling directions on the sphere and placing the camera along the corresponding ray toward the sphere center, assuming the camera always faces the center. The camera distance is determined from a reference distance $d$ that ensures all subject Gaussians tightly lie within the view, and the final camera centers are placed at random distances within $[\alpha d, \beta d]$ (\ie, $\alpha=1.1, \beta=1.6$) along each sampled ray. To avoid viewpoints from below the scene, the elevation angle on the sphere is restricted to $[-70^\circ, 90^\circ]$, and directions outside this range are not sampled.

For fair comparison, the number of sampled viewpoints for SVS is set to $\frac{3L(L-1)}{2}$, the same as in IVS. For both methods, the final camera pose is selected from the generated candidates as the one achieving the highest text-image alignment score measured by a VLM.

Visualizations of camera poses generated by IVS and SVS are shown in~\cref{fig:ivs_svs}. For reference, we present the rendered views with the top-4 alignment scores computed by FG-CLIP~\cite{xie2025fg}. IVS performs well when the training set already contains viewpoints aligned with the instruction, as interpolation can further refine such poses. In contrast, SVS is more effective when the instruction mainly specifies coarse spatial layouts (\eg, \textit{center}, \textit{left}, \textit{right}) without requiring specific camera orientations or detailed compositions. Additional comparisons between IVS, SVS and CapFrame are provided in~\cref{more_exp}.

\subsection{Estimating the Gravity Direction}
Our orientation loss includes a gravity term that encourages the camera up direction to align with the world up direction (Eq.~(19) in the main paper). We estimate this direction from the training images using COLMAP~\cite{schonberger2016structure}. Specifically, we apply COLMAP's \texttt{model\_orientation\_aligner}, which detects dominant vanishing directions under the Manhattan world assumption and aligns the reconstruction so that the $+Y$ axis corresponds to gravity. We then use this aligned $-Y$ axis as the world up direction in the \textit{gravity} loss.

\begin{figure}[htbp]
  \centering
  \includegraphics[width=\linewidth]{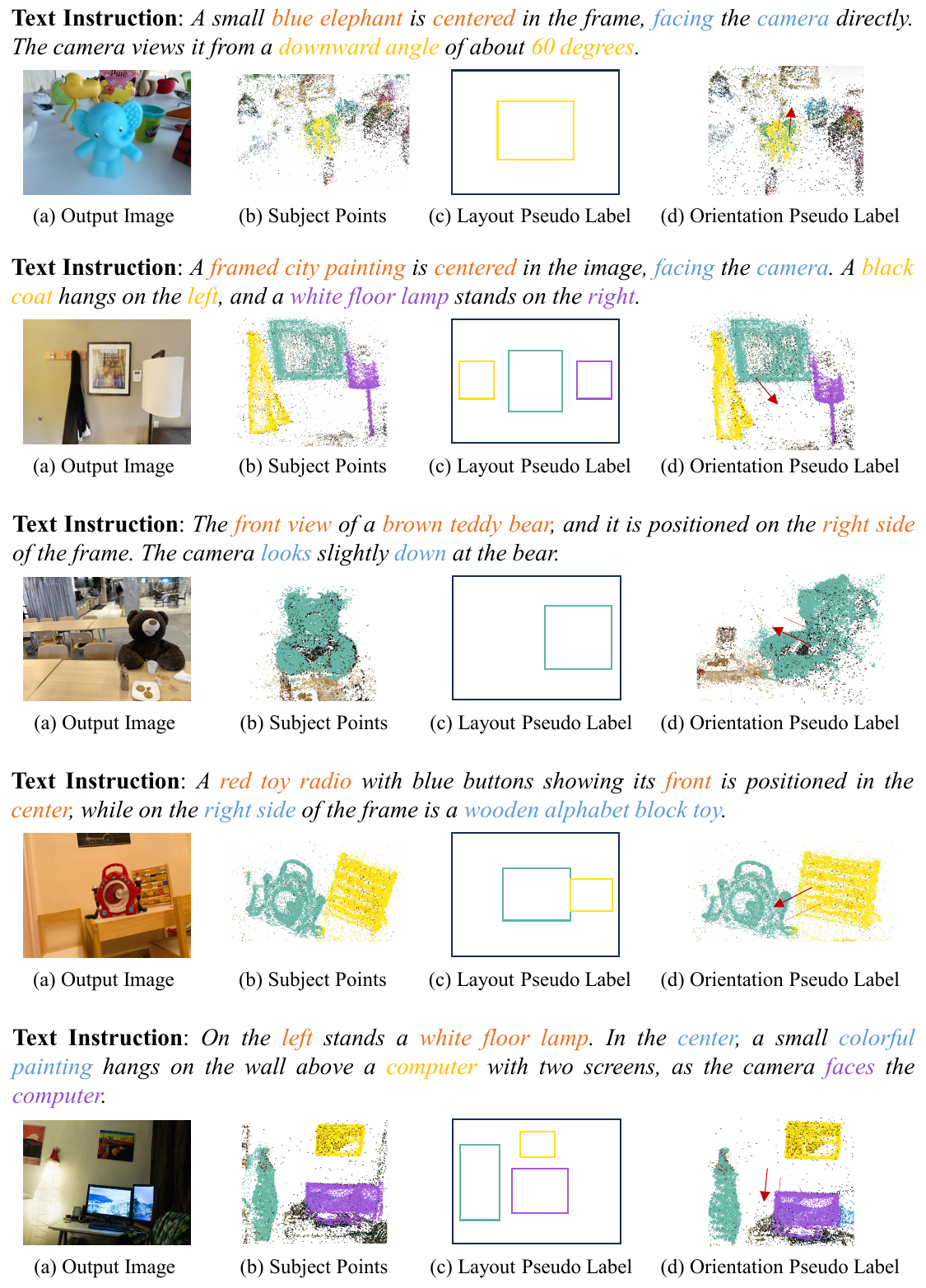}
  \caption{\textbf{Pseudo labels produced in the \textbf{Translate} stage.} For each instruction, (a) shows the final image rendered by CapFrame, (b) shows the 3D points / Gaussian subsets associated with the parsed subjects, (c) shows the layout pseudo labels as target 2D bounding boxes, and (d) shows the orientation pseudo label as a red ray originating from the subject mean center. Orientation pseudo labels are generated only for asymmetric subjects when the instruction specifies a viewing direction.}
  \label{fig:intermediate}
\end{figure}

\section{Visualization of Pseudo Labels}
\label{pseudo}
\Cref{fig:intermediate} illustrates the intermediate outputs produced in the \textbf{Translate} stage. This stage converts the input text instruction into geometric pseudo labels that guide the subsequent camera pose optimization.

Specifically, \cref{fig:intermediate} visualizes four components for each example: (a) the image rendered by CapFrame, (b) the 3D Gaussian subsets associated with the parsed subjects, (c) the layout pseudo labels represented as 2D bounding boxes in normalized image coordinates, and (d) the orientation pseudo labels describing the desired viewing direction.

For each subject mentioned in the instruction, the MLLM first extracts key entities and predicts layout and orientation cues. Grounded SAM~\cite{kirillov2023segment, liu2024grounding, ren2024grounded} is then applied to the retrieved views to obtain segmentation masks, whose pixels are back-projected into 3D using the rendered depth of the 3D Gaussian scene. The KNN search between these points and Gaussian means identifies the subset of Gaussians corresponding to each subject. 

Layout pseudo labels are inferred by the MLLM and represented as bounding boxes in normalized image coordinates, defining the target regions where subjects should appear in the rendered frame. During optimization, the associated Gaussians are projected onto the image plane to enforce these spatial constraints. Orientation pseudo labels are generated only for asymmetric subjects when the instruction specifies viewpoint cues (\eg, \textit{front view} or \textit{side view}). We rely on the MLLM to identify asymmetric subjects by reasoning about whether each object possesses a canonical facing direction. In such cases, the MLLM predicts orientation offsets that are converted into geometric targets for camera pose optimization. Otherwise, only layout pseudo labels are used.

The examples in~\cref{fig:intermediate} show cases in which both layout and orientation pseudo labels are applied, illustrating how compositional text instructions are translated into geometric constraints. The prompts used for the MLLM are provided in~\cref{prompt}.

\section{Computational Time}
\label{time}
The computational cost of CapFrame mainly comes from the \textbf{Retrieve}, \textbf{Translate}, and \textbf{Refine} stages. In our experiments, the number of top-$K$ views is fixed to $K=2$, so the cost of the \textbf{Retrieve} stage is dominated by similarity search over the training images.

The runtime of the \textbf{Translate} stage mainly scales with the number of parsed subjects, since subject extraction and Gaussian association are performed for each subject. The runtime of the \textbf{Refine} stage is largely determined by the image resolution and the number of optimization iterations.

We report the approximate runtime per scene in~\cref{tab:time}. Since early stopping is applied when the loss variation falls below a predefined threshold, optimization does not always reach the maximum of 1500 iterations. Moreover, because the initial viewpoint is randomly selected from the top-$K$ retrieved views, the number of iterations required for convergence varies across runs. Therefore, the reported runtimes should be interpreted as indicative estimates. In practice, the entire pipeline typically finishes within a few minutes.
\\\\
\noindent\textbf{Acceleration Strategy.} Our experiments span 38 scenes from 5 datasets, covering indoor, outdoor, tabletop and room-scale settings, with 12–411 input views and scene scales ranging from 12.3 to 311.8 in 3DGS coordinate units. Across these scenes, the average wall-clock time per query is 102.5\,s, comprising Retrieve (33.7\,s), Translate (25.3\,s), and Refine (43.5\,s). Since the \textbf{Refine} stage optimizes the camera pose using the coarse layout and spatial orientation of key subjects rather than fine pixel-level appearance, it is insensitive to rendering resolution and can be performed at reduced resolution during optimization. We therefore optimize the camera pose at low resolution and render the final frame at full resolution. This reduces the per-query runtime to 55.8\,s while incurring only minor deviations from the full-resolution pose ($\Delta R = 2.63^\circ$, $\Delta t = 0.17$).

\begin{table}[htbp]
\centering
\caption{\textbf{Per-scene runtime of CapFrame.} The first three columns report the dataset, scene and image resolution. \textbf{Stage 1+2} denotes the combined runtime of the \textbf{Retrieve} and \textbf{Translate} stages, \textbf{Stage 3} denotes the optimization time of the \textbf{Refine} stage, and \textbf{Total Time} reports the overall runtime. All values are in seconds.}
\resizebox{\linewidth}{!}{
\begin{tabular}{l|c|c|c|c|c}
\toprule
Dataset & Scene & Image Resolution & Stage 1+2 & Stage 3 & Total Time\\
\midrule

\multirow{5}{*}{Mip-NeRF}
    & Garden & $5187 \times 3361$ & 129.077 & 54.904 & 183.981 \\
    & Kitchen  & $3115 \times 2078$ & 89.373 & 252.599 & 341.973 \\
    & Room & $3114 \times 2075$ & 89.621 & 46.739 & 136.360 \\
    & Counter & $3115 \times 2076$ & 70.162 & 43.230 & 113.392 \\
    & Bonsai & $3118 \times 2078$ & 95.816 & 26.613 & 122.429 \\
\cline{1-6}

\multirow{13}{*}{Deep Blending}
    & Playroom & $1264 \times 832$ & 54.575 & 7.883 & 62.458 \\
    & Drjohnson  & $1332 \times 876$ & 61.229 & 22.329 & 83.558 \\
    & Tree & $2648 \times 1739$ & 28.461 & 54.582 & 83.043 \\
    & Library & $3569 \times 1987$ & 87.277 & 49.234 & 136.512 \\
    & Hugo & $3217 \times 2135$ & 50.428 & 123.592 & 174.020 \\
    & Creepyattic & $1231 \times 819$ & 74.183 & 7.393 & 81.576 \\
    & Bedroom & $1298 \times 840$ & 53.486 & 36.311 & 89.798 \\
    & Aquarium & $2642 \times 1962$ & 35.982 & 36.005 & 71.987 \\
    & Street & $2606 \times 1734$ & 31.709 & 53.470 & 85.179 \\
    & Yellowhouse & $4185 \times 2505$ & 31.979 & 139.556 & 171.536 \\
    & Ponche & $2503 \times 1466$ & 43.735 & 26.553 & 70.288 \\
    & Museum1 & $2338 \times 1537$ & 31.596 & 17.490 & 49.086 \\
    & Museum2 & $2353 \times 1538$ & 31.295 & 10.884 & 42.179 \\
\cline{1-6}

\multirow{4}{*}{Tanks}
    & Family & $960 \times 540$ & 46.690 & 13.159 & 59.848 \\
    & Ignatius  & $960 \times 540$ & 43.493 & 8.403 & 51.896 \\
    & Museum & $960 \times 540$ & 37.886 & 5.238 & 43.124 \\
    & Francis & $960 \times 540$ & 34.099 & 50.676 & 84.775 \\
\cline{1-6}

\multirow{4}{*}{LERF-OVS}
    & Teatime & $988 \times 730$ & 46.465 & 63.109 & 109.574 \\
    & Figurines  & $986 \times 728$ & 64.398 & 45.905 & 110.303 \\
    & Waldo Kitchen & $985 \times 725$ & 39.464 & 38.478 & 77.942 \\
    & Ramen & $988 \times 731$ & 36.334 & 10.825 & 47.160 \\
\cline{1-6}

\multirow{12}{*}{DL3DV-10K}
    & 1 & $976 \times 541$ & 60.229 & 39.321 & 99.550 \\
    & 2  & $953 \times 536$ & 56.859 & 13.969 & 70.828 \\
    & 3 & $954 \times 536$ & 63.523 & 37.702 & 101.225 \\
    & 4 & $975 \times 541$ & 58.062 & 21.856 & 79.918 \\
    & 5 & $955 \times 536$ & 61.067 & 10.845 & 71.912 \\
    & 6 & $958 \times 538$ & 50.637 & 18.851 & 69.488 \\
    & 7 & $963 \times 539$ & 60.711 & 58.378 & 119.089 \\
    & 8 & $3827 \times 2154$ & 199.226 & 178.983 & 378.209 \\
    & 9 & $976 \times 542$ & 58.173 & 34.229 & 92.402 \\
    & 10 & $964 \times 540$ & 62.860 & 26.259 & 89.118 \\
    & 11 & $960 \times 539$ & 66.426 & 11.164 & 77.590 \\
    & 12 & $958 \times 538$ & 71.609 & 33.517 & 105.126 \\

\bottomrule
\end{tabular}}
\label{tab:time}
\end{table}

\section{Additional Experimental Results}
\label{more_exp}
\begin{figure}[htbp]
  \centering
  \includegraphics[width=\linewidth]{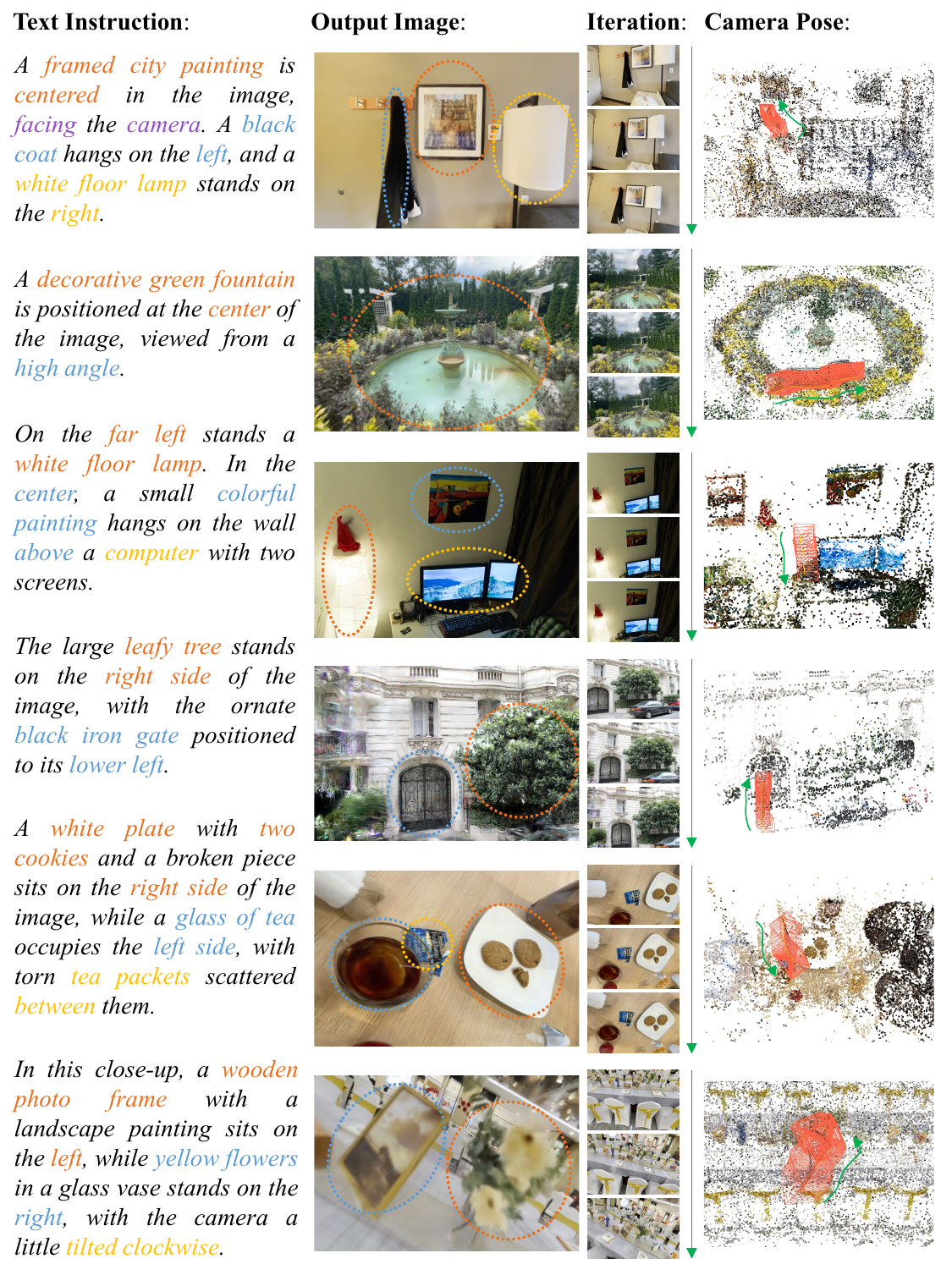}
\caption{\textbf{Additional text-instructed viewpoint grounding results.} Each row shows a text instruction, the final rendered image produced by CapFrame, intermediate rendered views during optimization, and the corresponding camera pose trajectory. CapFrame can satisfy both composition and viewing angle constraints from free-form compositional instructions.}
  \label{fig:more_capture}
\end{figure}

\begin{figure}[tb]
  \centering
  \includegraphics[width=\linewidth]{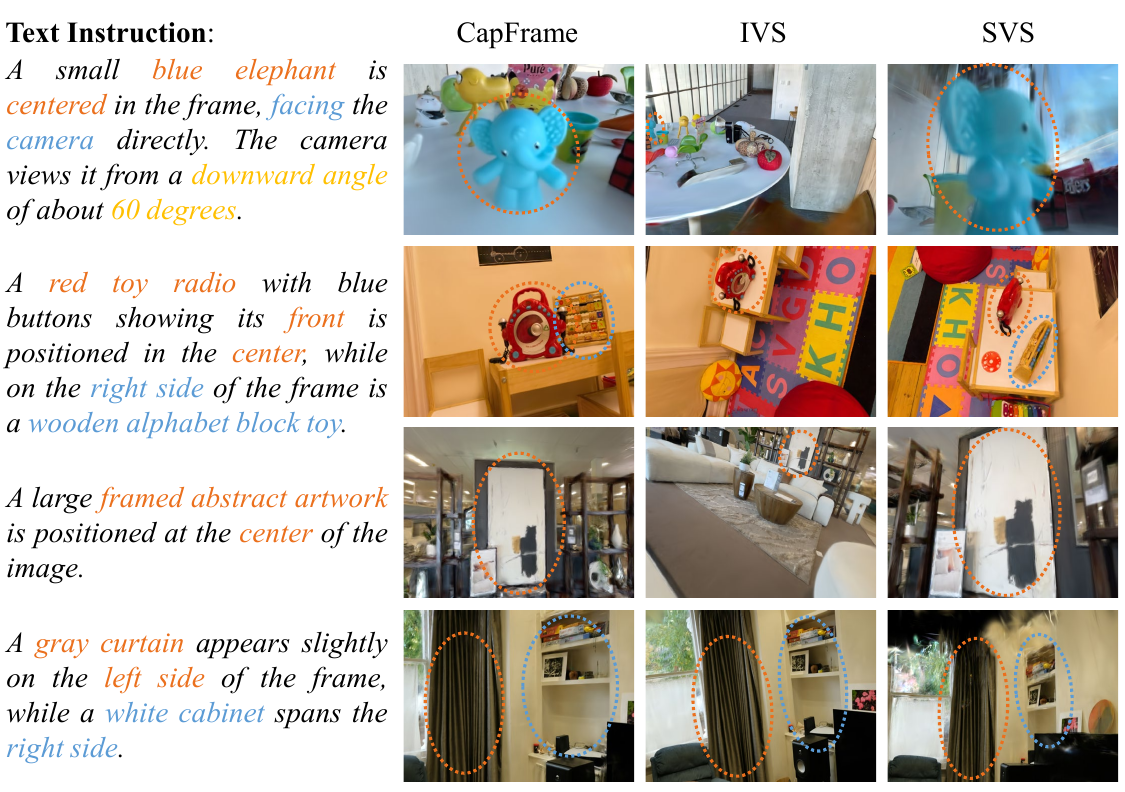}
  \caption{\textbf{Additional comparisons with baselines.} Each row corresponds to one text instruction, and the columns show the outputs of CapFrame, IVS and SVS. CapFrame more reliably satisfies joint layout and orientation constraints, especially when the desired viewpoint is weakly covered by the training cameras.}
  \label{fig:more_qualitative}
\end{figure}
\subsection{More Qualitative Results}
\noindent\textbf{More Results on Text-Instructed Viewpoint Grounding.}
Additional qualitative results are presented in~\cref{fig:more_capture}. 
CapFrame supports free-form text instructions and identifies camera viewpoints that satisfy both the specified composition and viewing angle by grounding the relevant subjects in the 3D Gaussian scene.
\\\\
\noindent\textbf{More Comparisons with IVS and SVS.}
Additional visual comparisons with IVS and SVS are shown in~\cref{fig:more_qualitative}. 
CapFrame consistently produces viewpoints aligned with the text instructions and avoids unnatural camera rotations, as illustrated in the \textit{red toy radio}. 
It also provides finer control over object orientation and camera viewing direction compared with the baselines, as demonstrated in the \textit{blue elephant}.

For IVS, when the training set already contains viewpoints consistent with the instruction, interpolation can refine such poses to produce aligned viewpoints, as seen in the \textit{gray curtain and white cabinet}. 
However, when the desired viewpoint lies outside the coverage of the training views, interpolation fails to locate a consistent solution. For SVS, when instructions specify simple spatial attributes such as \textit{center}, sampling-based exploration can identify reasonable viewpoints, as demonstrated in the \textit{framed abstract artwork}. However, when the instruction involves more complex layouts and viewing angles, both IVS and SVS struggle to produce suitable viewpoints. 
This limitation is compounded by the limited perception capability of FG-CLIP~\cite{xie2025fg}, which is used to score candidate views.

\begin{figure}[htbp]
  \centering
  \includegraphics[width=\linewidth]{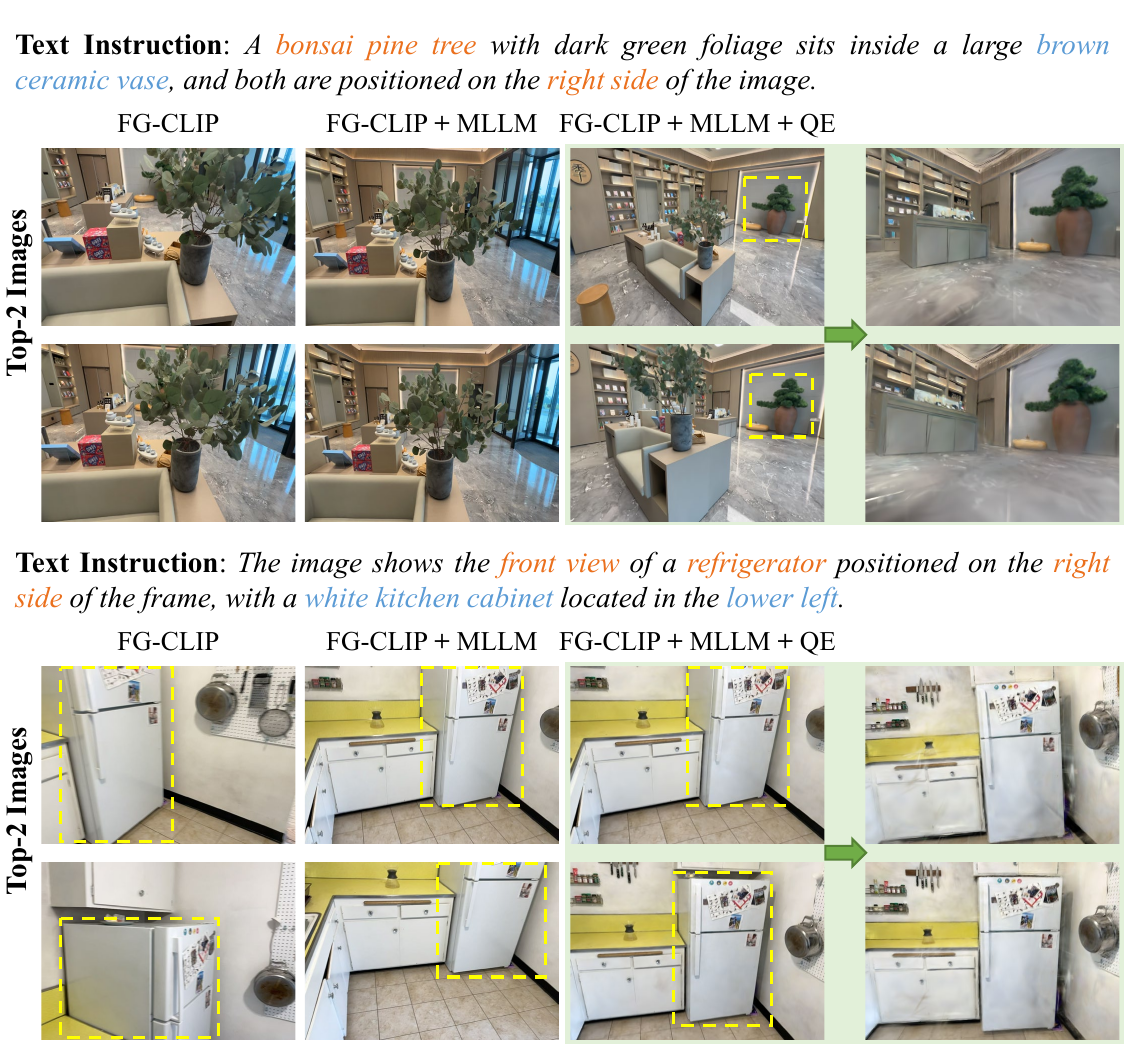}
  \caption{\textbf{Additional ablation of initialization.} For each instruction, we show the top-2 retrieved views obtained by FG-CLIP, FG-CLIP+MLLM, and FG-CLIP+MLLM+QE, together with the final optimized results. QE better disambiguates visually similar objects and photographic attributes, leading to stronger initial views for subsequent optimization.}
  \label{fig:more_abla_init}
\end{figure}

\subsection{More Component Analyses}
\begin{figure}[h]
  \centering
  \includegraphics[width=\linewidth]{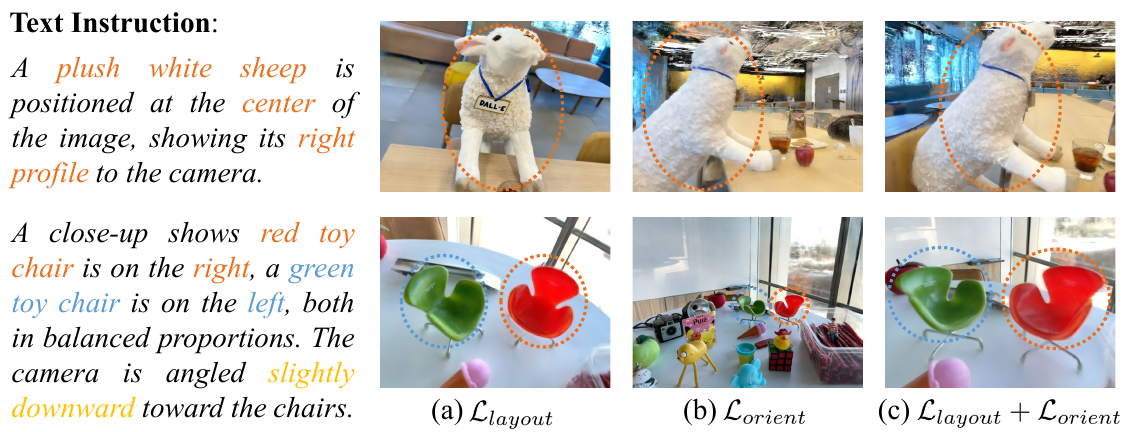}
  \caption{\textbf{Additional ablation of loss.} For each instruction, (a) uses only $\mathcal{L}_{\text{layout}}$, (b) uses only $\mathcal{L}_{\text{orient}}$, and (c) uses both. Using both losses is necessary to jointly control subject placement, scale and viewing direction.}
  \label{fig:more_abla_loss}
\end{figure}

We present additional component analyses of the \textit{Question-Evaluation} (QE) strategy in the \textbf{Retrieve} stage and the layout and orientation losses in the \textbf{Refine} stage, as illustrated in~\cref{fig:more_abla_init} and~\cref{fig:more_abla_loss}.
\\\\
\noindent\textbf{Question-Evaluation Strategy.} The QE strategy decomposes the instruction into a set of evaluation questions using an MLLM. These questions assess key photographic aspects such as subject visibility, frame-based layout, camera-based orientation, and overall text-image alignment. 
For each candidate view, the same MLLM evaluates the image by answering these questions and assigning scores. 
The final score of each view is obtained by aggregating the MLLM-provided scores across all questions. 
The prompts used for question generation and evaluation are provided in~\cref{prompt}.

\Cref{fig:more_abla_init} highlights the advantage of QE. In cluttered scenes containing visually similar objects, the \textit{bonsai pine tree} in the background can be easily confused with a similar potted plant in the foreground. Without QE, the background subject is difficult to distinguish, causing the \textit{bonsai pine tree} to be overlooked and leading to unsuccessful optimization. 
Moreover, QE guides the MLLM to attend to photographic attributes in the instruction. As shown in the \textit{refrigerator} example, QE enables the MLLM to retrieve training images that better satisfy textual cues such as \textit{front view} and \textit{right side}, providing a stronger initialization for subsequent optimization.
\\\\
\noindent\textbf{Layout and Orientation Losses.}
\Cref{fig:more_abla_loss} demonstrates the role of the layout and orientation losses. 
Without the orientation loss, the \textit{white sheep} fails to present its right profile toward the camera. Removing the layout loss instead causes the \textit{white sheep} to shift away from the intended central position. 
A similar effect is observed in the \textit{chairs} example: without the orientation loss, the camera fails to achieve the required slight downward viewing angle, while removing the layout loss makes it difficult to control the scale of the \textit{chairs} in the frame and their distance from the camera.

\subsection{Beyond Source-View Initialization}
\begin{figure}[htbp]
  \centering
  \includegraphics[width=\linewidth]{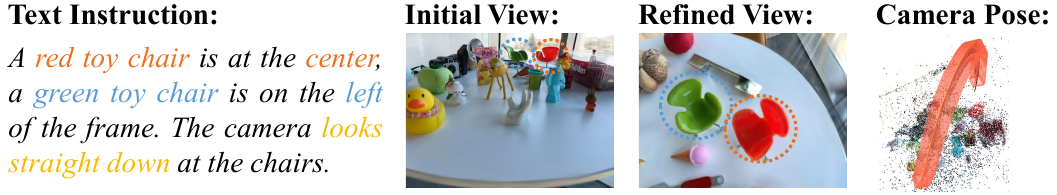}
  \caption{\textbf{Refined view deviates from initial view.} The pose trajectory (right) illustrates the substantial change from initial view to the refined view.}
  \label{fig:beyond_src}
\end{figure}

CapFrame does not assume that the final camera lies close to a source view. It only requires an initialization in which the target subject is visible, so that the subject Gaussians and their differentiable masks can be obtained. Then, the optimization proceeds in continuous $SE(3)$. As a result, CapFrame can discover novel viewpoints absent from the input image set. In~\cref{fig:beyond_src}, we deliberately initialize from a source view distant from the desired camera, yet CapFrame still recovers a bird's-eye-like view unseen among the source images, with the refined pose lying far from the initial one ($\Delta R = 57.76^\circ$, $\Delta t = 4.285$ in 3DGS coordinate units). Such behavior illustrates CapFrame's ability to discover viewpoints beyond the source views.

\section{Failure Cases}
\label{fail}

\begin{figure}[htb]
  \centering
  \includegraphics[width=\linewidth]{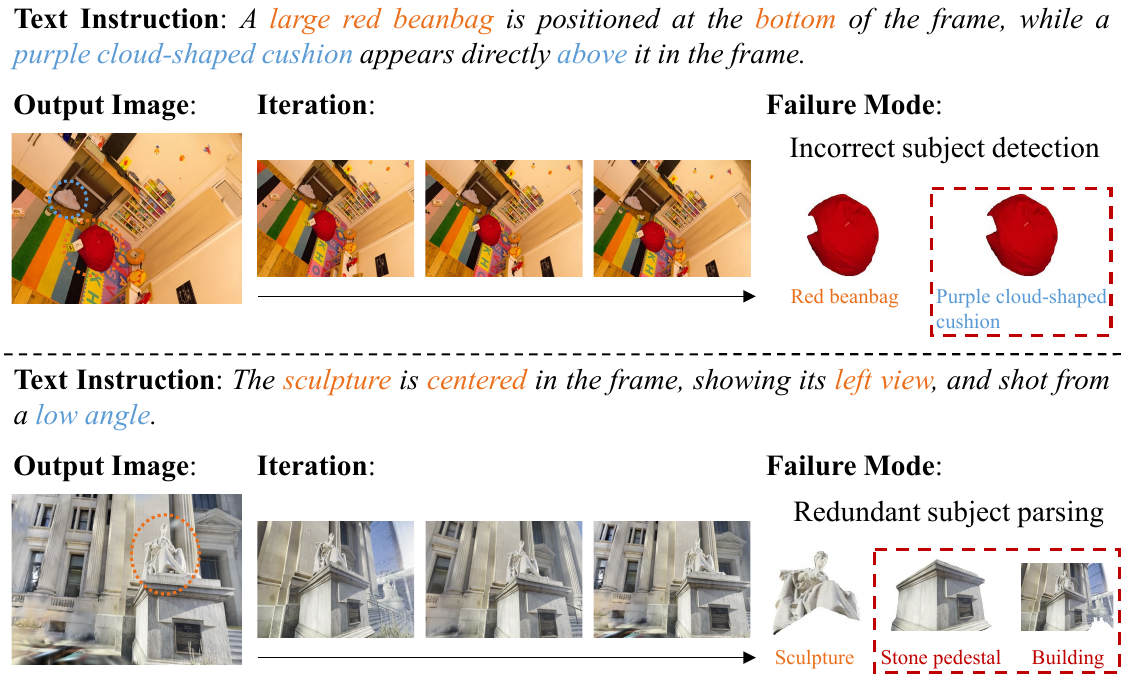}
  \caption{\textbf{Failure cases.} For each example, the left shows the final output, the middle shows intermediate rendered views during optimization, and the right highlights the source of failure. The top row shows incorrect subject detection, where Grounded SAM~\cite{ren2024grounded} localizes the wrong object. The bottom row shows redundant subject parsing, where the MLLM~\cite{bai2025qwen3} introduces irrelevant objects that add noisy layout constraints.}
  \label{fig:failure}
\end{figure}

CapFrame mainly exhibits two failure modes.
\\\\\noindent\textbf{Incorrect Subject Detection.}
The first failure mode originates from the subject localization step in the \textbf{Translate} stage. 
Grounded SAM~\cite{ren2024grounded} is used to segment subjects in the retrieved views before associating them with 3D Gaussians. 
Due to the limitations of open-vocabulary perception, incorrect detections may occur. 
As shown in~\cref{fig:failure}, the detector fails to correctly identify the \textit{purple cloud-shaped cushion}. 
Consequently, the Gaussian subset associated with the intended subject is incorrect, leading to inaccurate layout supervision. 
Although the \textit{red beanbag} gradually moves toward the bottom region during optimization, the rendered result cannot satisfy the text instruction.
\\\\\noindent\textbf{Redundant Subject Parsing.}
The second failure mode arises from the subject parsing process in the \textbf{Translate} stage. 
When decomposing the instruction, the MLLM~\cite{bai2025qwen3} may occasionally identify visually salient but textually irrelevant objects (\eg, \textit{stone pedestal} and \textit{building}) as additional subjects. 
For such objects, the instruction provides no layout or orientation cues, yet layout pseudo labels are still generated during the translation process. 
These additional layout constraints introduce noisy supervision in the layout loss, which interferes with camera pose optimization and may lead to suboptimal viewpoints.

\begin{figure}[htbp]
  \centering
  \includegraphics[width=\linewidth]{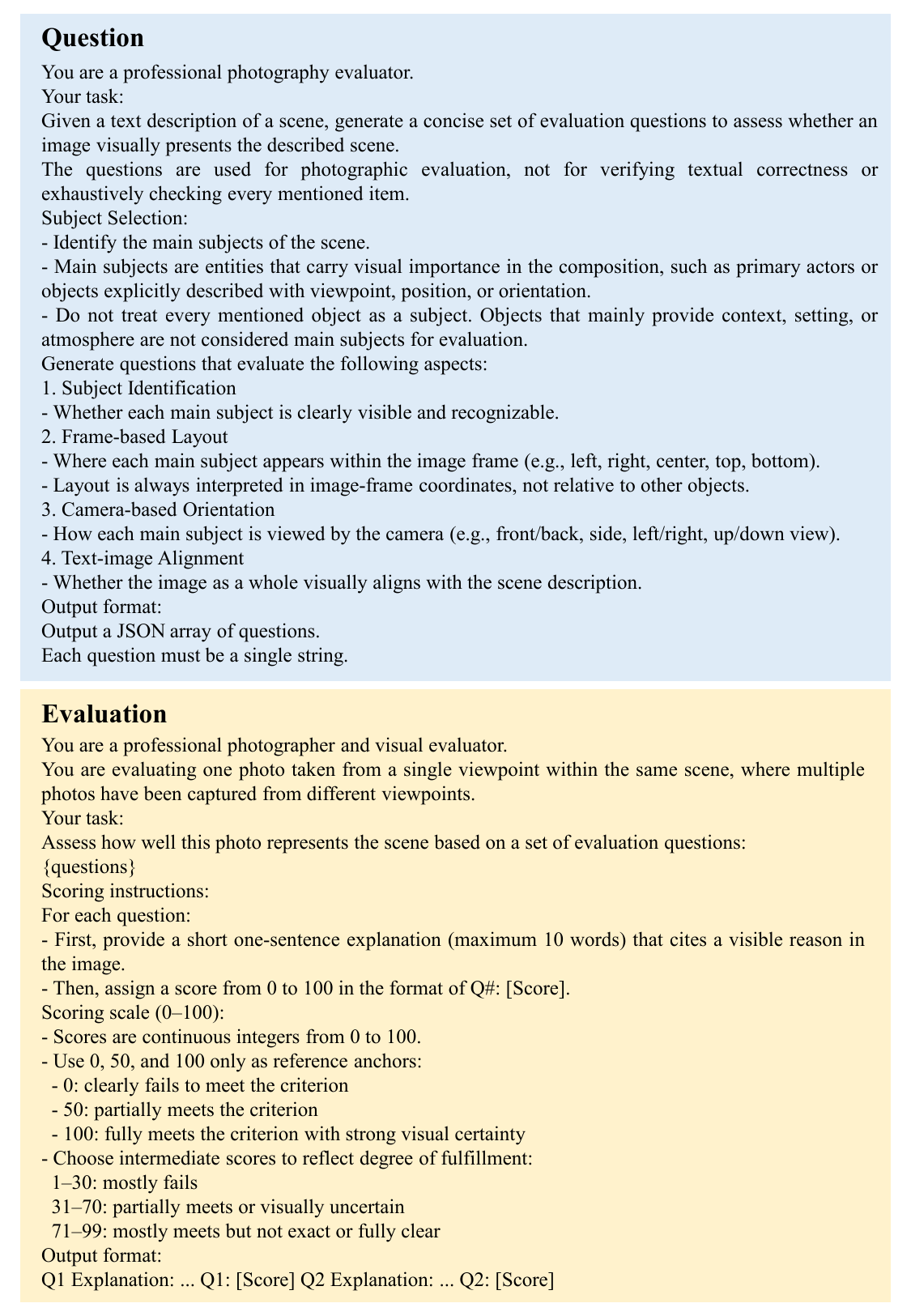}
  \caption{\textbf{Prompts used in the \textbf{Retrieve} stage.} The top prompt asks the MLLM to generate evaluation questions from the text instruction, and the bottom prompt asks the MLLM to score a candidate image by answering those questions.}
  \label{fig:qe_prompt}
\end{figure}

\begin{figure}[htbp]
  \centering
  \includegraphics[width=\linewidth]{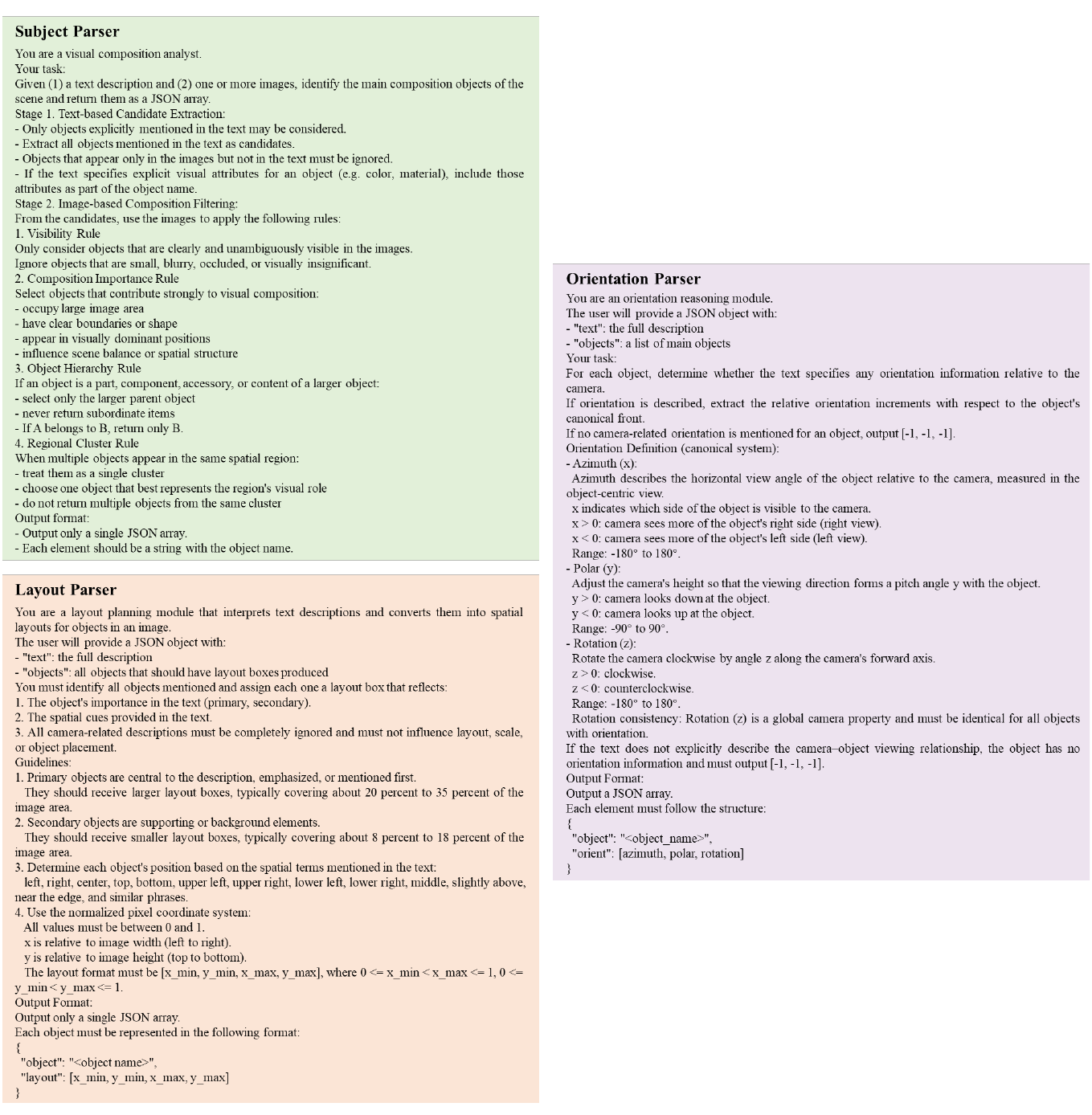}
  \caption{\textbf{Prompts used in the \textbf{Translate} stage.} From left to right, we show the prompts for the subject parser, the layout parser, and the orientation parser, which convert a free-form instruction into geometric pseudo labels for camera pose optimization.}
  \label{fig:trans_prompt}
\end{figure}

\section{Prompts for MLLM}
\label{prompt}

The MLLM is used in both the \textbf{Retrieve} and \textbf{Translate} stages.

In the \textbf{Retrieve} stage, we employ the Question-Evaluation (QE) strategy to assess candidate views from multiple photographic perspectives. As shown in~\cref{fig:qe_prompt}, the MLLM first generates a set of evaluation questions from the text instruction, focusing on four aspects: subject identification, frame-based layout, camera-based orientation, and overall text-image alignment. Each candidate image is then evaluated by the same MLLM, which provides a short explanation and a score for each question. The final score of each view is obtained by aggregating the per-question scores.

In the \textbf{Translate} stage, the MLLM decomposes the instruction into structured photographic attributes. Specifically, three types of prompts are used, as illustrated in~\cref{fig:trans_prompt}: (1) a \textit{subject parser} that extracts the main subjects mentioned in the instruction, (2) a \textit{layout parser} that predicts the desired spatial arrangement of these subjects within the frame, and (3) an \textit{orientation parser} that infers camera viewing angles or subject-facing directions when such cues are present in the instruction. These outputs are subsequently converted into geometric pseudo labels that guide camera pose optimization.

\begin{figure}[htbp]
  \centering
  \includegraphics[width=0.9\linewidth]{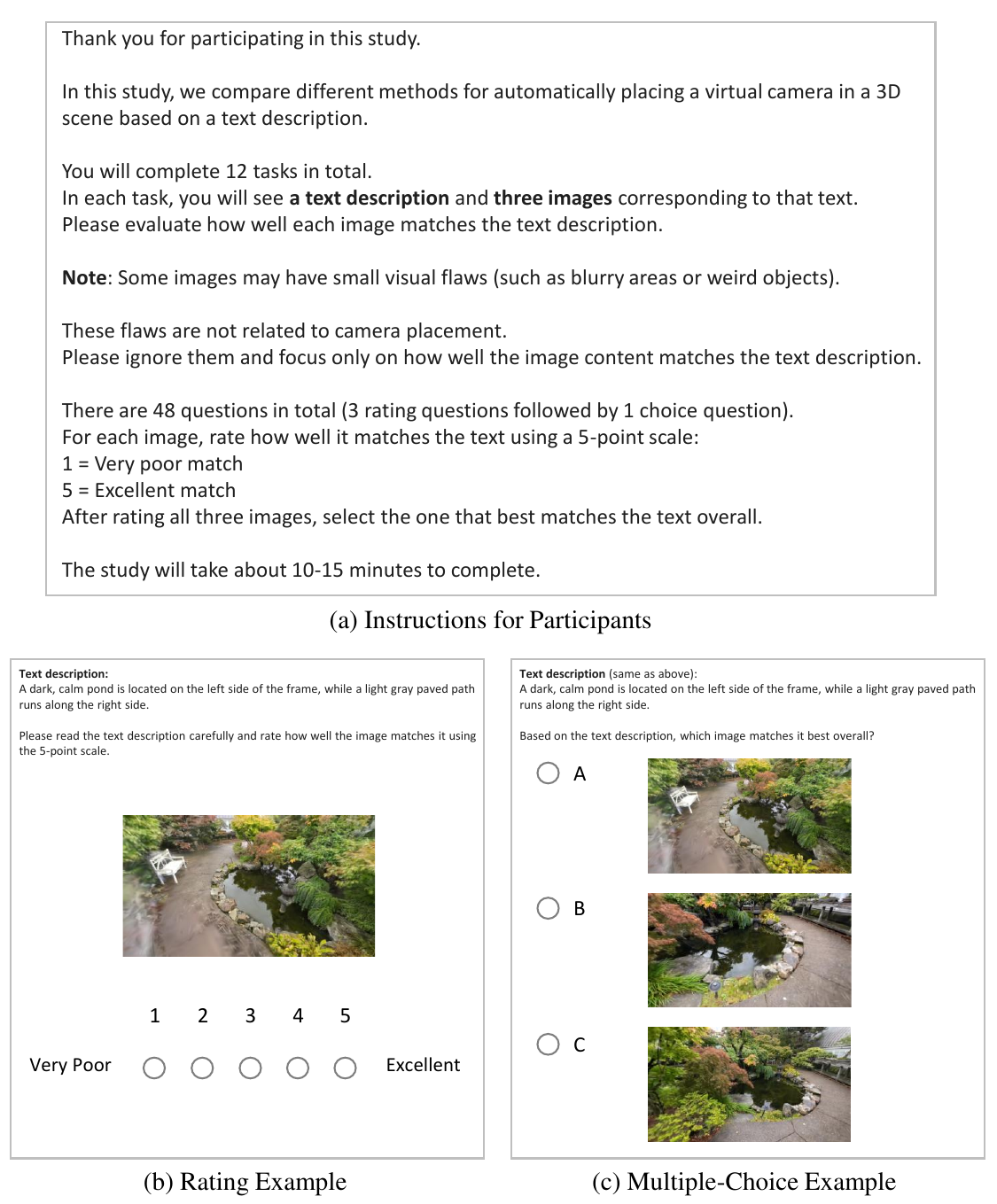}
  \caption{\textbf{Example user-study interface.} (a) Instructions shown to participants. (b) Example of the 5-point rating question. (c) Example of the final multiple-choice question used to select the best-matching image.}
  \label{fig:user_study}
\end{figure}

\section{User Study Details}
\label{user_study}
A total of 33 participants took part in the user study, including individuals both with and without computer science backgrounds. The study evaluates how well rendered images align with the given text instructions.

The evaluation consists of 12 question groups. In each group, participants are presented with one text instruction and three candidate images generated by different methods. Participants first rate each image on a 5-point scale according to how well the visual content matches the instruction (1: very poor match; 5: excellent match). After rating all three images, they select the image that best matches the instruction overall.

In total, the study contains 48 questions: three rating questions and one selection question per group. The order of the candidate methods is randomized to avoid presentation bias. Participants are instructed to focus only on the alignment between the image content and the text description, while disregarding visual artifacts unrelated to camera placement. Examples of the questionnaire are shown in~\cref{fig:user_study}.

\end{document}